\pdfoutput=1

\documentclass[11pt]{article}

\usepackage{ACL2023}

\usepackage{times}
\usepackage{latexsym}

\usepackage[T1]{fontenc}

\usepackage[utf8]{inputenc}

\usepackage{microtype}

\usepackage{inconsolata}

\usepackage{graphicx}
\usepackage{subcaption}
\usepackage{booktabs} 
\usepackage{amsmath}
\usepackage{multirow}
\usepackage{soul}

\usepackage{hyperref}  
\usepackage{appendix}

\usepackage[most]{tcolorbox}

\lstdefinestyle{pythonstyle}{
    language=Python,
    basicstyle=\ttfamily\small,
    breaklines=true,
    morecomment=[l][\color{gray}]{\#},
    keywordstyle=\color{blue},
    stringstyle=\color{green!50!black},
    commentstyle=\color{gray},
    showstringspaces=false,
    numbers=left,
    numberstyle=\tiny\color{gray},
    numbersep=5pt,
    frame=single,
    framesep=5pt,
    xleftmargin=15pt,
    tabsize=4,
}

\title{From Calculation to Adjudication: Examining LLM Judges on Mathematical Reasoning Tasks}

\author{Andreas Stephan\textsuperscript{1,2}, Dawei Zhu\textsuperscript{4}, Matthias Aßenmacher\textsuperscript{6,7}, \\ \bf Xiaoyu Shen\textsuperscript{5},  Benjamin Roth\textsuperscript{1,3} \\
\textsuperscript{1}Faculty of Computer Science, 
\textsuperscript{2}UniVie Doctoral School Computer Science, \\
\textsuperscript{3}Faculty of Philological and Cultural Studies, University of Vienna, Vienna, Austria \\
\textsuperscript{4}Saarland University, Saarland Informatics Campus, \textsuperscript{5}Eastern Institute of Technology, Ningbo \\
\textsuperscript{6}Department of Statistics, LMU Munich, \textsuperscript{7}Munich Center for Machine Learning (MCML) \\
 \small{
   \textbf{Correspondence:} \href{mailto:andreas.stephan@univie.ac.at}{andreas.stephan@univie.ac.at}
 }
}

\begin{document}

\maketitle
\begin{abstract}
To reduce the need for human annotations, large language models (LLMs) have been proposed as judges of the quality of other candidate models.
The performance of LLM judges is typically evaluated by measuring the correlation with human judgments on generative tasks such as summarization or machine translation.
In contrast, we study LLM judges on mathematical reasoning tasks. These tasks require multi-step reasoning, and the correctness of their solutions is verifiable, enabling a more objective evaluation.
We perform a detailed performance analysis and find that easy samples are easy to judge, and difficult samples are difficult to judge.
Our analysis uncovers a strong correlation between judgment performance and the candidate model task performance, indicating that judges tend to favor higher-quality models even if their answer is incorrect. 
As a consequence, we test whether we can predict the behavior of LLM judges using simple features such as part-of-speech tags and find that we can correctly predict 70\%-75\% of judgments. 
We conclude this study by analyzing practical use cases, showing that LLM judges consistently detect the on-average better model but largely fail if we use them to improve task performance.
\footnote{Code will be made available upon acceptance.}
\end{abstract}


\section{Introduction}

The automatic evaluation of machine learning models promises to reduce the need for human annotations. Specifically, the LLM-as-a-judge paradigm \cite{zheng2023judging} has gained traction, aiming to assess or compare the quality of generated texts automatically. This approach is beneficial for automated data labeling \cite{tan2024largelanguagemodelsdata}, self-improvement of LLMs \cite{wu2024metarewardinglanguagemodelsselfimproving}, and ranking the capabilities of LLMs, potentially concerning specific tasks \cite{zheng2023judging}.
\begin{figure}
    \centering
    \includegraphics[width=0.93\linewidth]{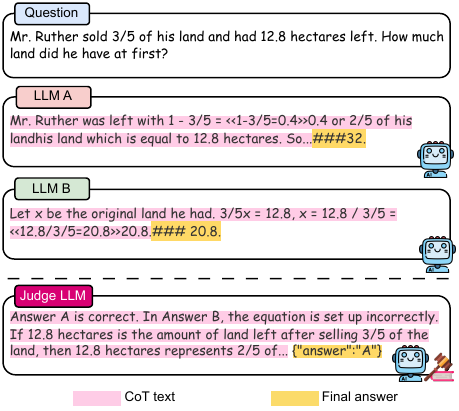}
    \caption{In our problem setup two LLMs ($A$ and $B$), 
    provide candidate answers for a math problem, and a judge LLM \textit{has} to decide which one is correct. All three use chain-of-thought (CoT) reasoning \cite{wei2022chain}.}
    \label{fig:figure1}
\end{figure}
Much like judges in the real world, who are expected to be exact, fair, and unbiased 
\cite{bangalore_principles}, LLM judges, should be unbiased and logical.
Previous works investigate properties and biases of LLM judges on generative tasks such as translation or summarization \cite{kim2024biggen, liu2024llmsnarcissisticevaluatorsego}, typically evaluated using correlation with human annotators, and thus being inherently subjective.

In this work, we investigate LLM judges on mathematical reasoning datasets. Such tasks require complex multi-step reasoning and judgments can be analyzed through the lense of verifiable solutions, allowing us to investigate the relationship between judge and candidate models in a principled manner.
In our setup, LLM Judges are given two answers and they have to classify whether both answers, one of them (which one), or none is correct (see Figure \ref{fig:figure1}.
We base our analysis on four large ($>27B$ parameters) LLMs and four small ($<10B$ parameters) LLMs on three mathematical reasoning datasets.

Our experiments contain a detailed performance examination. We find that the best tested judge is LLama 3.1 70B, reaching 60\% to 90\% judgment performance, depending on the dataset. Our results confirm the intuition that judgment performance is aligned with task difficulty.

We perform a statistical analysis of judgment performance and model quality. We find that the individual task performances of judge and candidate models are highly indicative features of judgment performance, as they explain most of the variance in a linear model (as measured by $R^2$).
On the subset of questions where the candidate models give one correct and one incorrect answer, we uncover an intriguing correlation between judge performance and candidate models' task performance, indicating that LLM judges tend to select incorrect answers from better models. 

We hypothesize that judges partially base their judgment on linguistic cues rather than solely on the reasoning withinin the answers.
We follow literature analysing machine-generated text \cite{shaib-etal-2024-detection} and find that 70\%-75\% of the judgments can be predicted using simple linguistic features, highlighting the systematicity behind the judge decisions.

Lastly, we analyze practical use cases and discuss usage recommendations. Our experiments suggest that LLM judges reliably detect the model of higher task performance but can not reliably improve task performance. Rather, we find that it is more sensible to use the judge model as an answer generator, and subsequently take the majority vote of all three answers.


In summary, our contributions are as follows:

\begin{enumerate}
\item We perform an in-depth performance analysis of LLM judges on three diverse mathematical reasoning tasks.
\item We identify a correlation between model quality, as measured by task performance, and judgment performance, indicating that LLM judges are biased towards higher-quality models.
\item We are able to predict the LLM judgment with 70\%-75\ accuracy using only stylistic patterns, e.g. N-grams of POS-tags. This indicates that LLMs, to a large degree, judge independently of the reasoning.
\item We find that judges reliably detect the model of higher quality but are not able to reliably improve task performance.
\end{enumerate}

\newpage
\section{Related Work}

\subsection{LLM as Judges}

Using \emph{LLMs as judges} to evaluate text generated by LLMs, including their own outputs, has recently attracted significant interest because it reduces the need for human annotation \cite{zheng2023judging}. Typically, large state-of-the-art models are used as judges. Applications include the automatic assessment of language model capabilities and, such as ranking models with respect to their competence on a given task \cite{zheng2023judging}, and reinforcement learning from AI feedback by automatically generating data for preference optimization \cite{bai2022constitutional, wu2024metarewardinglanguagemodelsselfimproving}.

Various methods exist to make judgments \cite{zheng2023judging, liusie-etal-2024-llm}. One approach is pairwise selection \cite{wang2024pandalm}, where two answers are presented, and the model is asked to select the better one. Another approach is pointwise grading \cite{li2024generative}, where the model is asked to assign a grade based on a predefined scale, and the answer with a better grade is chosen. 
Judgment prompts may involve reference solutions or not.
Another body of research explicitly trains models to act as judges \cite{kim2024prometheus, wang2024pandalm} or closely related, as reward models \cite{wang2024helpsteer2, li2024generative}. 

The effectiveness of LLMs as judges is typically assessed by measuring the correlation or overlap with human judgments \cite{zheng2023judging, kim2024biggen}. In contrast, we focus on tasks with a concrete final answer. 
Finally, we want to stress that several works caution against the use of LLM judges as experts \cite{bavaresco2024llmsinsteadhumanjudges,  koo2023benchmarkingcognitivebiaseslarge, raina2024llmasajudge, doddapaneni-etal-2024-finding}. 

\subsection{Biases in LLM-as-a-judge}

Human-annotated data inherently reflects the annotators' biases and opinions. These biases can be detrimental or (intentionally) beneficial, depending on the goals of the annotation process \cite{plank-2022-problem}. Similarly, several studies have explored the biases present in LLM judges:

One linguistic bias is ordering bias \cite{zheng2023judging, koo2023benchmarkingcognitivebiaseslarge,wang-etal-2024-large-language-models-fair}, where a judge gives a different answer depending on the order in which answers are presented. \citet{panickssery2024llmevaluatorsrecognizefavor} note that it is possible to interpret position bias as a sign that the model is unsure. 
There are multiple works \cite{xu2024prideprejudicellmamplifies, panickssery2024llmevaluatorsrecognizefavor, liu2024llmsnarcissisticevaluatorsego} that find evidence for self-bias or self-preference.
\citet{koo2023benchmarkingcognitivebiaseslarge} provide a benchmark for analyzing cognitive biases.
\citet{west2024the} and \citet{oh-etal-2024-generative} explore the ``Generative AI Paradox'' where it is easier for LLMs to generate solutions rather than analyzing them, unlike humans who often find analysis easier than generation.

In this work, we aim to establish a better understanding of underlying patterns that relate judgments to interpretable factors, such as task performance or stylistic patterns.

\section{General Setup }
\label{sec:setup}

In the following, we describe the problem setting, including the used notation, and the general experimental setting including used models and datasets.

\subsection{Problem Description}

In this work, we use an LLM judge, referred to as $J$, to assess answers produced by two other candidate LLMs, $A$ and $B$, in response to math questions (see Figure \ref{fig:figure1} for an illustrative example). The two candidate answers may both be correct, both incorrect, or either the answer of model $A$ or $B$ correct. The judge's task is to determine which of these cases applies by reviewing both the CoT reasoning and the final responses provided in candidate answers. 

Thus, the judge engages in a four-class classification task. We denote the judge's accuracy by the score $S^{J}_{A,B}$ and call this metric \textit{judgment performance}.
Further, we define the \textit{task performance} of an individual model $X$ on a specific dataset as $S_X$, e.g. $S_A$, $S_B$ or $S_J$.



\subsection{Datasets}

The experiments encompass three mathematical reasoning datasets where models highly benefit from multi-step CoT reasoning. For all datasets, we use accuracy as the performance metric.\\
\textbf{AQUA-RAT} \cite{ling-etal-2017-program} is a dataset to test the quantitative reasoning ability of LLMs. Unlike the other two datasets, the questions are multiple-choice. 
\textbf{GSM8K} \cite{cobbe2021gsm8k} consists of grade school math word problems. The answers are free-form numbers. 
\textbf{MATH} \cite{hendrycksmath2021} contains challenging competition mathematics problems. Find more details in Appendix \ref{app_sec:datasets}

\subsection{Models}

We evaluate the performance of openly available LLMs, including four large models including \textit{Qwen 2.5 72B} \cite{qwen2}, \textit{Llama 3.1 70B} \cite{llama3modelcard}, \textit{Yi 1.5 34B} \cite{ai2024yi}, \textit{Mixtral 8x7B} \cite{jiang2024mixtralexperts} and four small models, namely \textit{Llama 3 8B} \cite{llama3modelcard}, \textit{Gemma 1.1 7B} \cite{gemmateam2024gemmaopenmodelsbased}, \textit{Mistral 7B v0.3} \cite{jiang2023mistral7b}, and \textit{Mistral 7B v0.1} \cite{jiang2023mistral7b}. We use the chat- or instruction-tuned model variants and test each model as a candidate answer generator and as a judge. More information is in Appendix \ref{app_subsec:models}.

\subsection{Text Generation}

This section describes the generation of candidate answers and judgments. Find more information on prompts and hardware details in Appendix \ref{app_sec:experimental_setup}.

\paragraph{Candidate answer generation.} For each model we sample two CoT solutions using 4-shot prompting by setting the temperature to $0.9$. By generating two answers $a_1, a_2$ from the same model, we can also evaluate judgments of two different answers by the same model.

\begin{table*}[t]
    \centering
\resizebox{.98\textwidth}{!}{
\begin{tabular}{ll|cccccccc}
\toprule
 &  & Llama 3.1 70B & Qwen 2.5 72B & Qwen 2.5 14B & Gemma 2 27B & Qwen 2.5 7B & Gemma 2 9B & Llama 3.1 8B & Gemma 2 2B \\
\midrule
\multirow[t]{3}{*}{(1) $S^J_{A,B}$} & GSM8K & \textbf{90.05} & 85.39 & \underline{89.2} & 81.96 & 81.92 & 83.60 & 79.96 & 64.33 \\
 & AQUA-RAT & \textbf{74.47} & 69.26 & \underline{72.26} & 68.48 & 65.09 & 67.48 & 66.26 & 60.97 \\
 & MATH & \underline{61.18} & 58.03 & \textbf{62.36} & 55.34 & 50.70 & 52.92 & 50.96 & 50.35 \\
\cline{1-10}
\multirow[t]{3}{*}{(2) Same answer} & GSM8K & \underline{95.27} & \textbf{95.46} & 95.02 & 92.27 & 92.86 & 94.70 & 88.13 & 75.50 \\
 & AQUA-RAT & \textbf{79.8} & 77.74 & 77.19 & \underline{78.67} & 77.21 & 77.94 & 74.52 & 76.01 \\
 & MATH & \textbf{79.09} & \underline{77.91} & 76.86 & 75.39 & 73.14 & 75.65 & 71.05 & 77.85 \\
\cline{1-10}
\multirow[t]{3}{*}{(3) Different Answer} & GSM8K & \textbf{70.04} & 55.67 & \underline{68.43} & 47.09 & 49.93 & 49.50 & 51.41 & 31.71 \\
 & AQUA-RAT & \underline{57.44} & 48.92 & \textbf{57.64} & 45.55 & 40.45 & 46.31 & 44.10 & 30.76 \\
 & MATH & \underline{48.95} & 45.40 & \textbf{51.47} & 42.66 & 36.70 & 38.86 & 36.41 & 32.50 \\
\cline{1-10}
\multirow[t]{3}{*}{(4) 1-correct} & GSM8K & \textbf{78.18} & 64.08 & \underline{76.92} & 52.25 & 56.19 & 58.10 & 59.26 & 27.78 \\
 & AQUA-RAT & \underline{66.43} & 57.10 & \textbf{69.22} & 44.44 & 47.13 & 54.43 & 52.93 & 24.05 \\
 & MATH & \underline{71.92} & 70.19 & \textbf{79.62} & 41.80 & 57.79 & 60.97 & 60.73 & 22.62
 \\
\cline{1-10}
\bottomrule
\end{tabular}
}
\caption{Performance of judge LLMs (1) on all samples, (2) on samples where $A$ and $B$ agree, (3) on samples where $A$ and $B$ disagree and (4) on samples where exactly one given answer is correct. Results are averaged over all candidate model pairs $(A, B)$. The highest accuracy is \textbf{bold} and the second highest \underline{underlined}.}
\label{tab:performance_per_dataset}
\end{table*}


\paragraph{Judgements.} For all $36$ unique model combinations $(A, B)$\footnote{We consider all pairs from the eight LLMs, including self-pairing, yielding $\binom{8+2-1}{2}=36$ combinations.}, each model as judge $J$ and each sample of a dataset, we generate a zero-shot judgment.
In the case of self-pairing, i.e., $A=B$, we use both generated candidate answers, $a_1$ and $a_2$. Otherwise, for consistency, we always use the same sampled candidate answer $a_1$.
We accommodate positional bias \cite{zheng2023judging, koo2023benchmarkingcognitivebiaseslarge} by prompting in both possible orders. We obtain the judgment performance by averaging how often the judgments were correctly classified across orderings.

\section{Performance Analysis}
\label{sec:general_performance}

The experiments have multiple degrees of freedom, such as judges, candidate models, and datasets. To gain a comprehensive understanding of judges' behavior, we consider two perspectives. First, we investigate judge performance for each dataset, aiming to associate judge performance with task difficulty. Second, for a fixed dataset, we analyze judge performance across different pairs of candidate models.

\subsection{General Performance}

First, we compare how often the judges make a correct classification across different datasets and different subsets of the datasets. 

\begin{figure}[t]  
\centering
\begin{subfigure}{0.263\textwidth}  
\centering
\includegraphics[width=\textwidth]{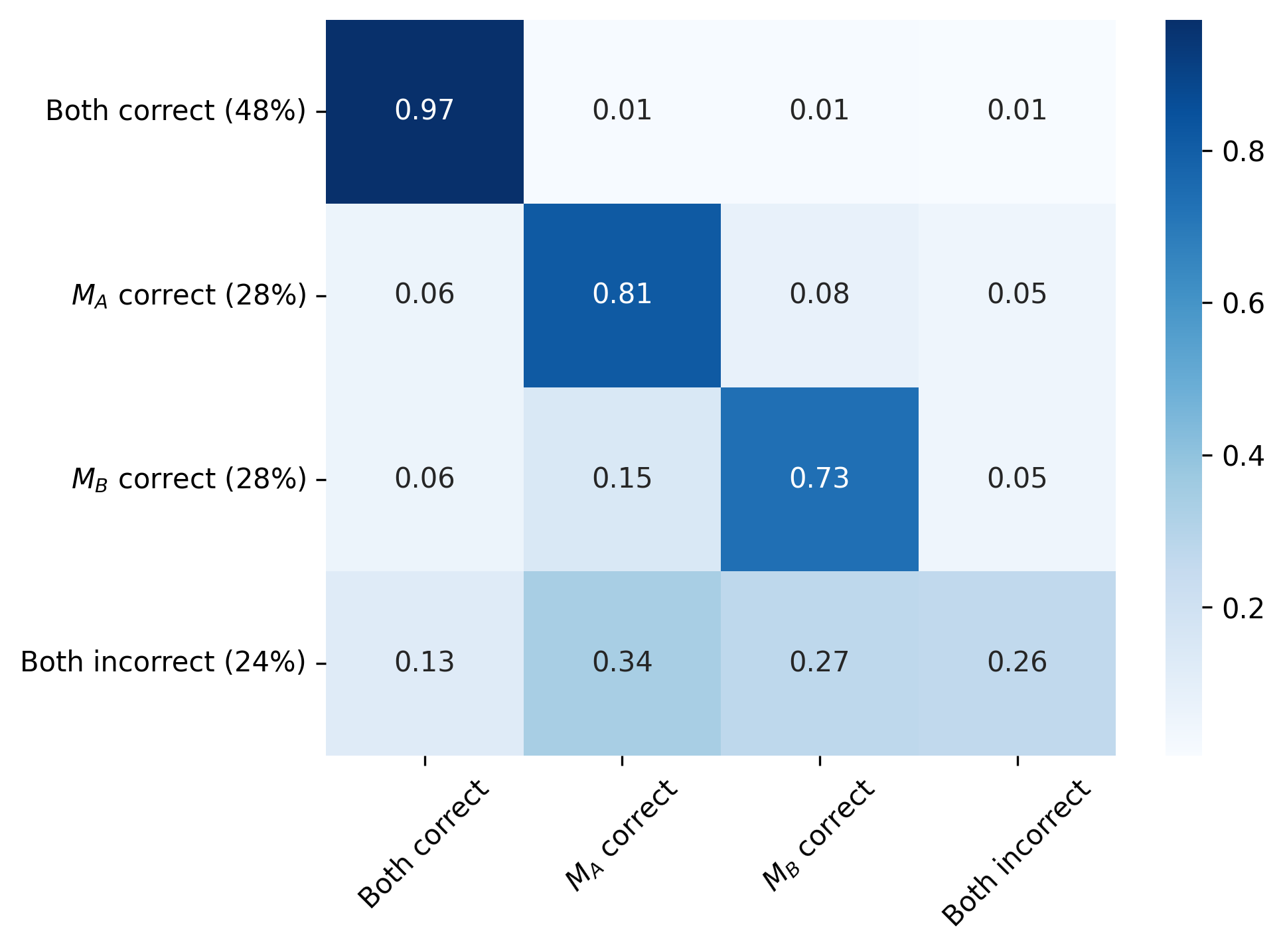}
\caption{Qwen2.5 72B}
\end{subfigure}
\begin{subfigure}{0.2\textwidth}
\centering
\includegraphics[width=\textwidth]{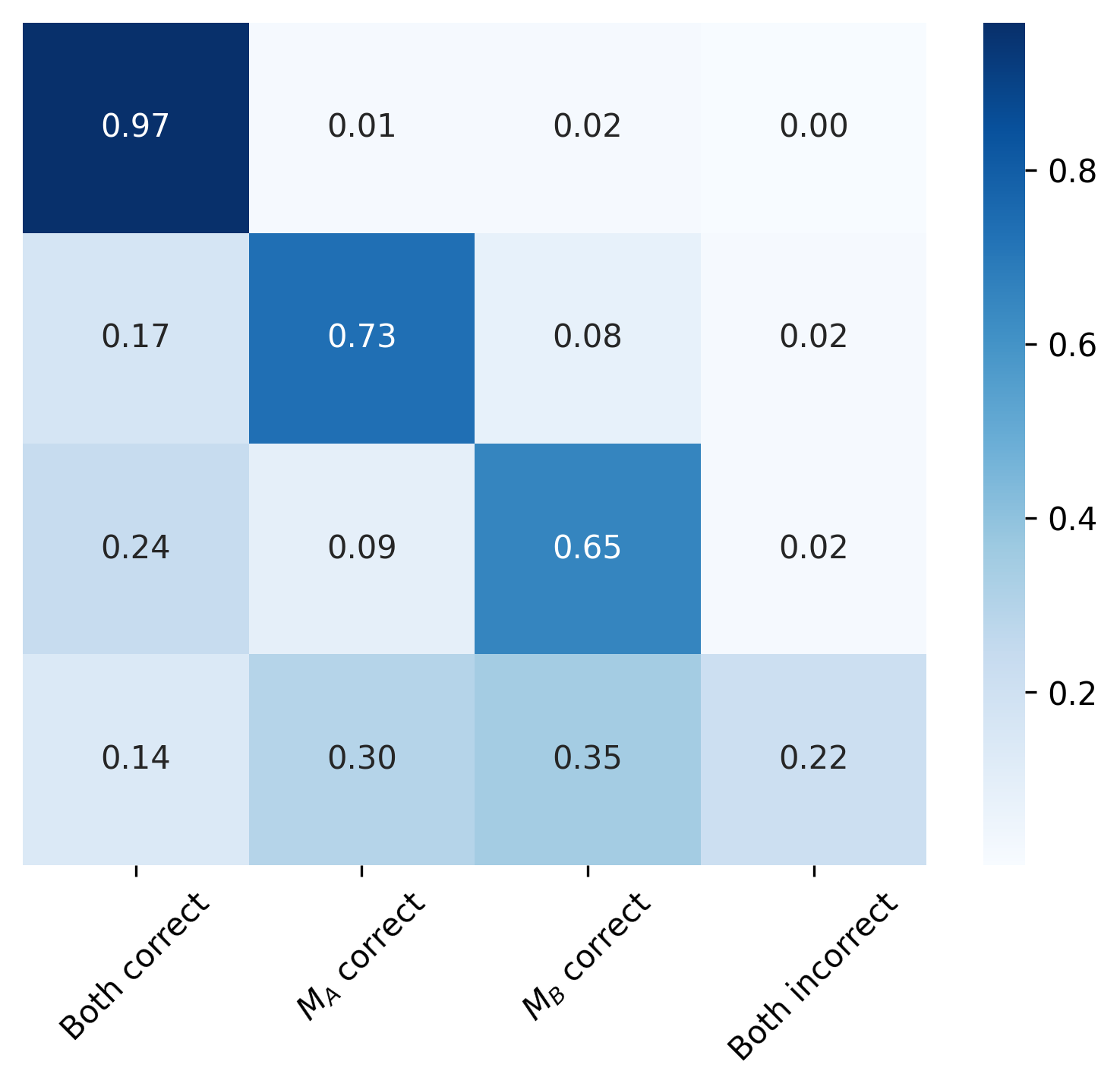}
\caption{Qwen 2.5 14B}
\end{subfigure}
\begin{subfigure}{0.262\textwidth}  
\centering
\includegraphics[width=\textwidth]{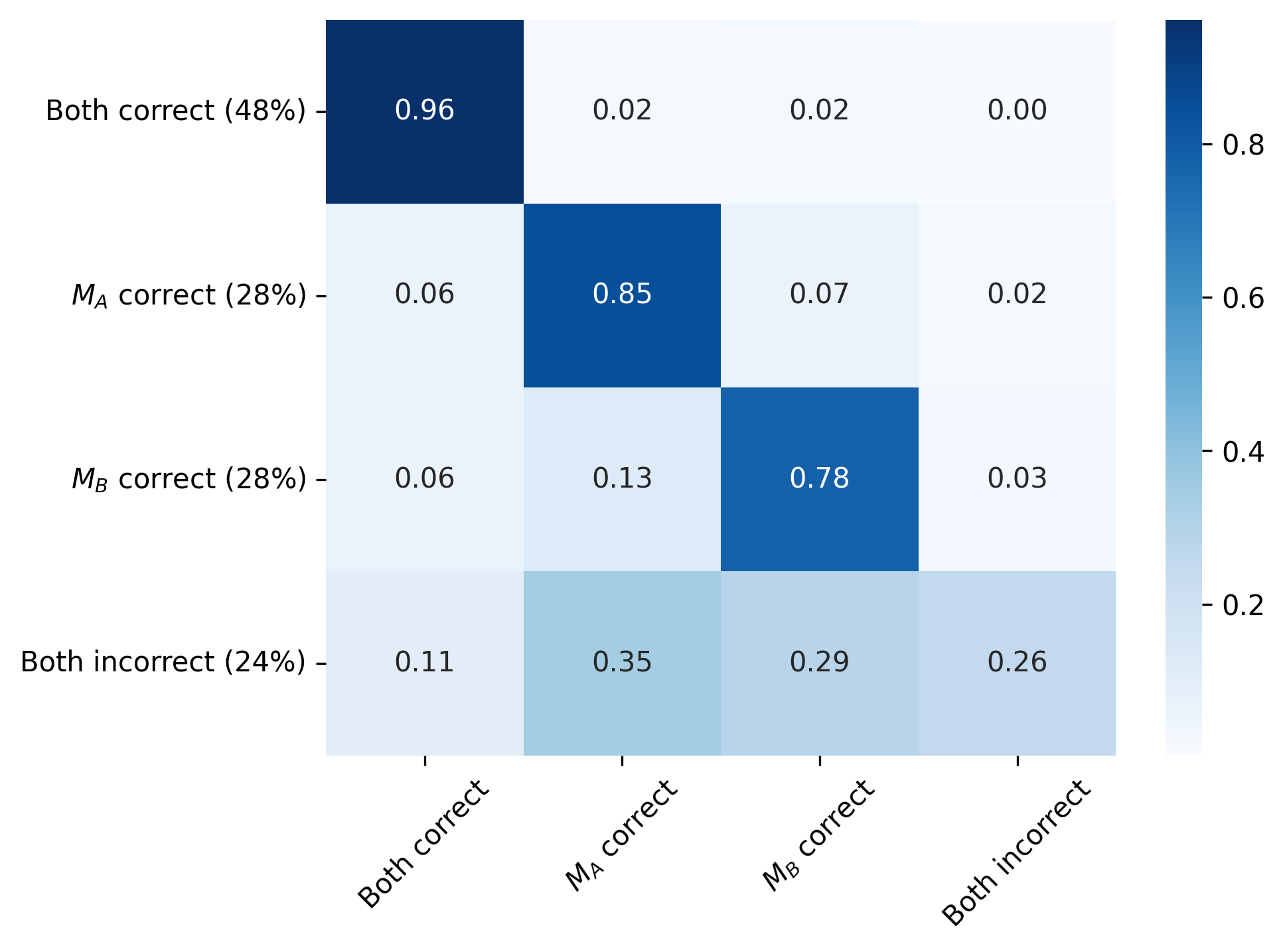}
\caption{Llama 3.1 70B}
\end{subfigure}
\begin{subfigure}{0.2\textwidth}  
\centering
\includegraphics[width=\textwidth]{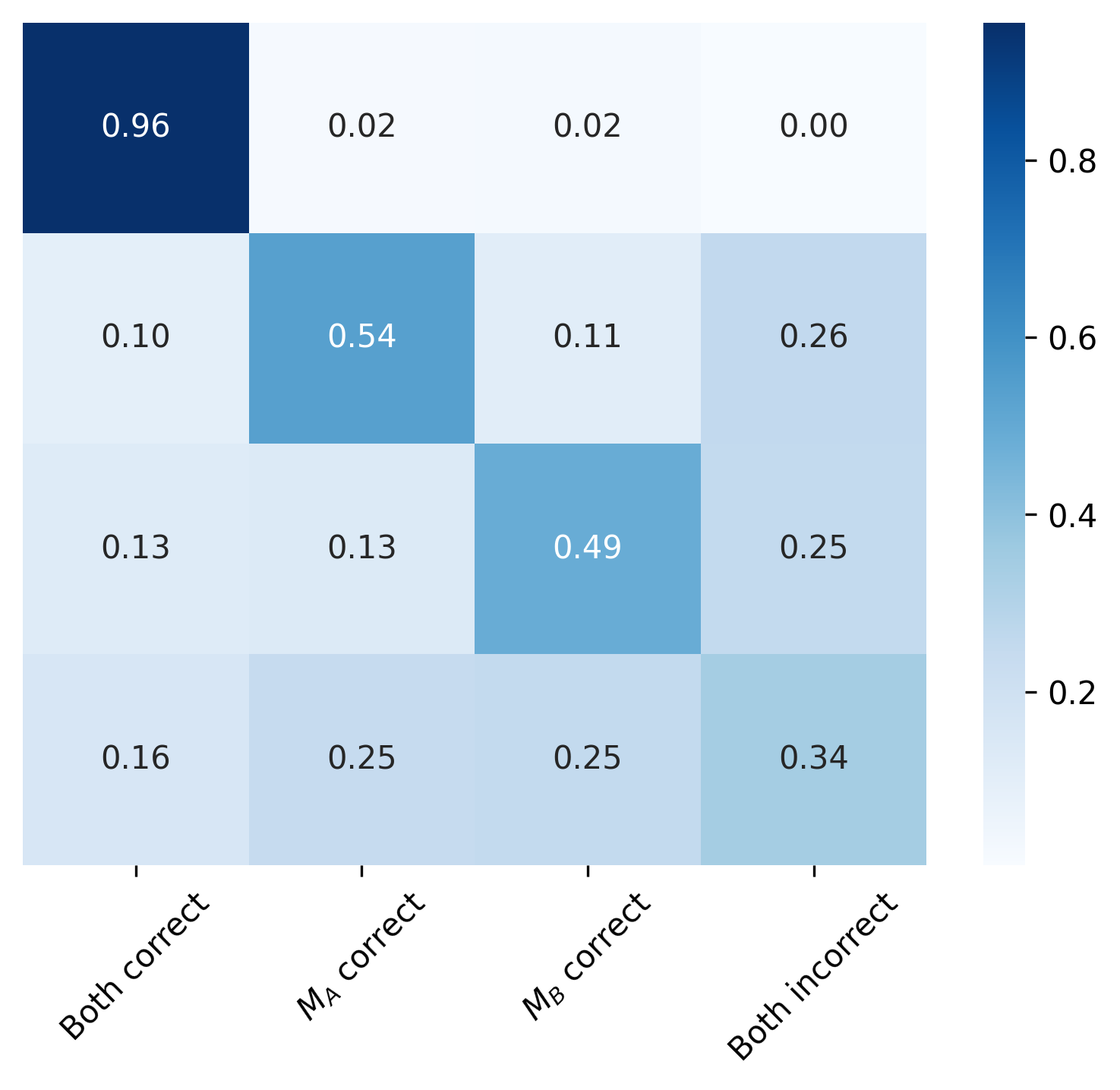}
\caption{Gemma 2 27B}
\end{subfigure}
\caption{Class confusion matrices per model. We observe that it is difficult for judges to detect that both answers are incorrect.}
\label{fig:confusion_matrices}  
\end{figure}

\paragraph{Setup.} We analyze multiple cases, each corresponding to a specific subset of the data.
\textit{Case (1)} investigates the observed judgment performance $S^J_{A,B}$ on the full dataset and \textit{Case (2)} analyzes the subset where both models give the same answer ($A=B$). \textit{Case (3)} shows the performance where both models give a different answer $(A \neq B$) and \textit{Case (4)} describes the performance on the subset where exactly one answer is correct. 
The results are shown in Table \ref{tab:performance_per_dataset}.
Further, we show the class confusion matrix for the four best-performing judges in Figure \ref{fig:confusion_matrices}.

\paragraph{Results.}
In general, we observe in Table \ref{tab:performance_per_dataset} that larger models outperform smaller models, with LLama 3.1 70B performing the best. 
Interestingly, Qwen 2.5 14B outperforms Qwen 2.5 72B. 
As shown in Figure \ref{fig:confusion_matrices}, the LLM judges have a performance of larger than 95\% if both answers are correct. Conversely, the most challenging situation is when both answers are incorrect. It seems that the difficulty of a problem also transcends the difficulty of making a judgment.
This is not necessarily intuitive. For instance, humans may find it easier to detect individual wrong reasoning steps and identify wrong answers, respectively.

In cases where one answer is correct and one answer is incorrect, we observe a moderate performance of the judges, reaching up to 80\% accuracy (see Case (4) in Table \ref{tab:performance_per_dataset} and Figure \ref{fig:confusion_matrices}). 

In Case (3) where both answers disagree, we observe moderate performance for large models of up to 70\%. Here, the smallest model Gemma 2 2B, has a low performance of around 35\%.
In what follows, we mostly focus on the analysis of the four largest LLMs as judges.


\subsection{Performance per model combination}

Each model has unique strengths and weaknesses and often answers different questions correctly. In this section, we analyze the judgment performance per model pair to gain a better understanding of the impact of candidate model combinations on judgment performance.

\begin{figure}
\centering
\begin{subfigure}{0.228\textwidth}
    \centering
    \includegraphics[width=\textwidth]{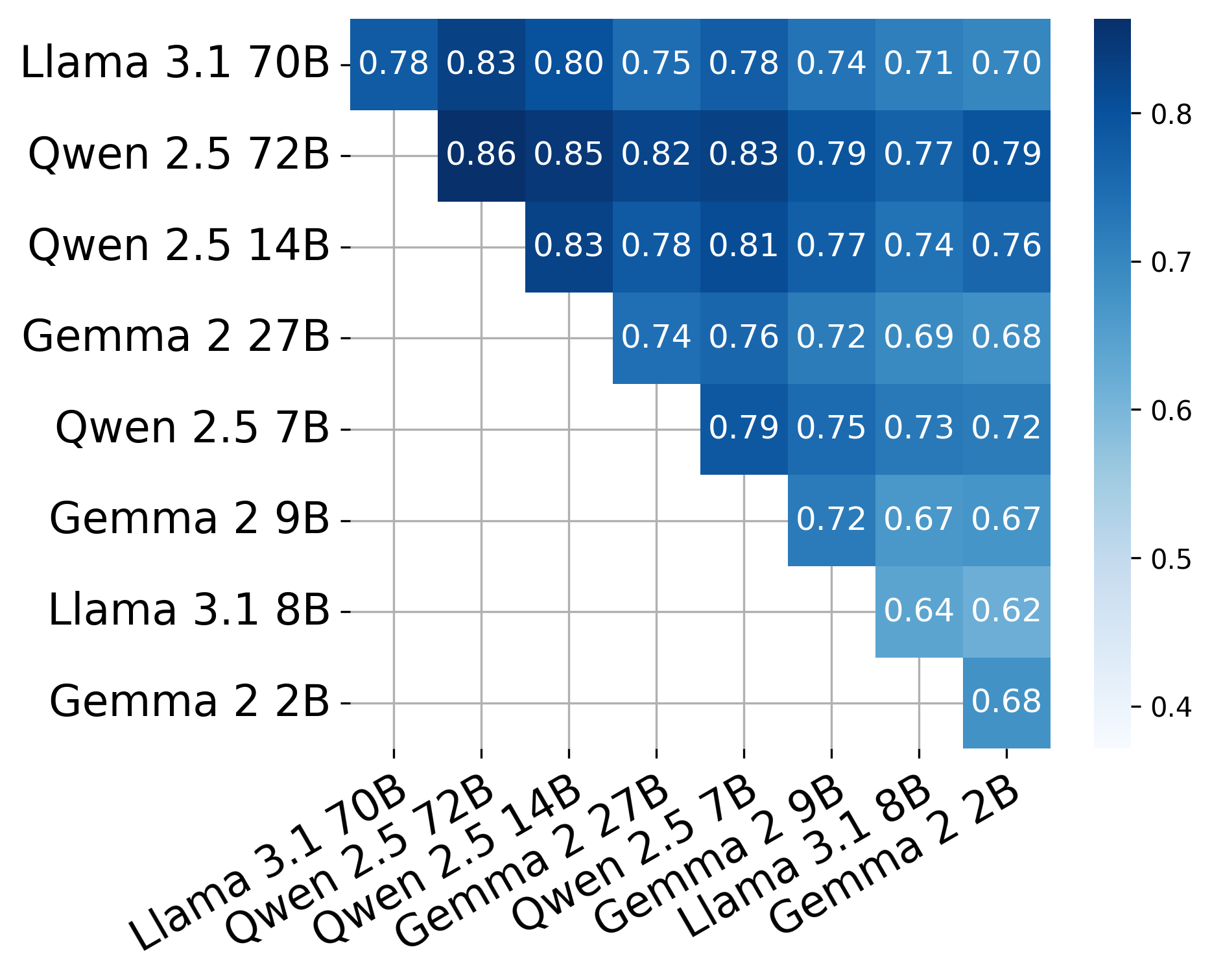}
    \caption{LLama 3.1 70B}
\end{subfigure}
\begin{subfigure}{0.235\textwidth}
    \centering
    \includegraphics[width=\textwidth]{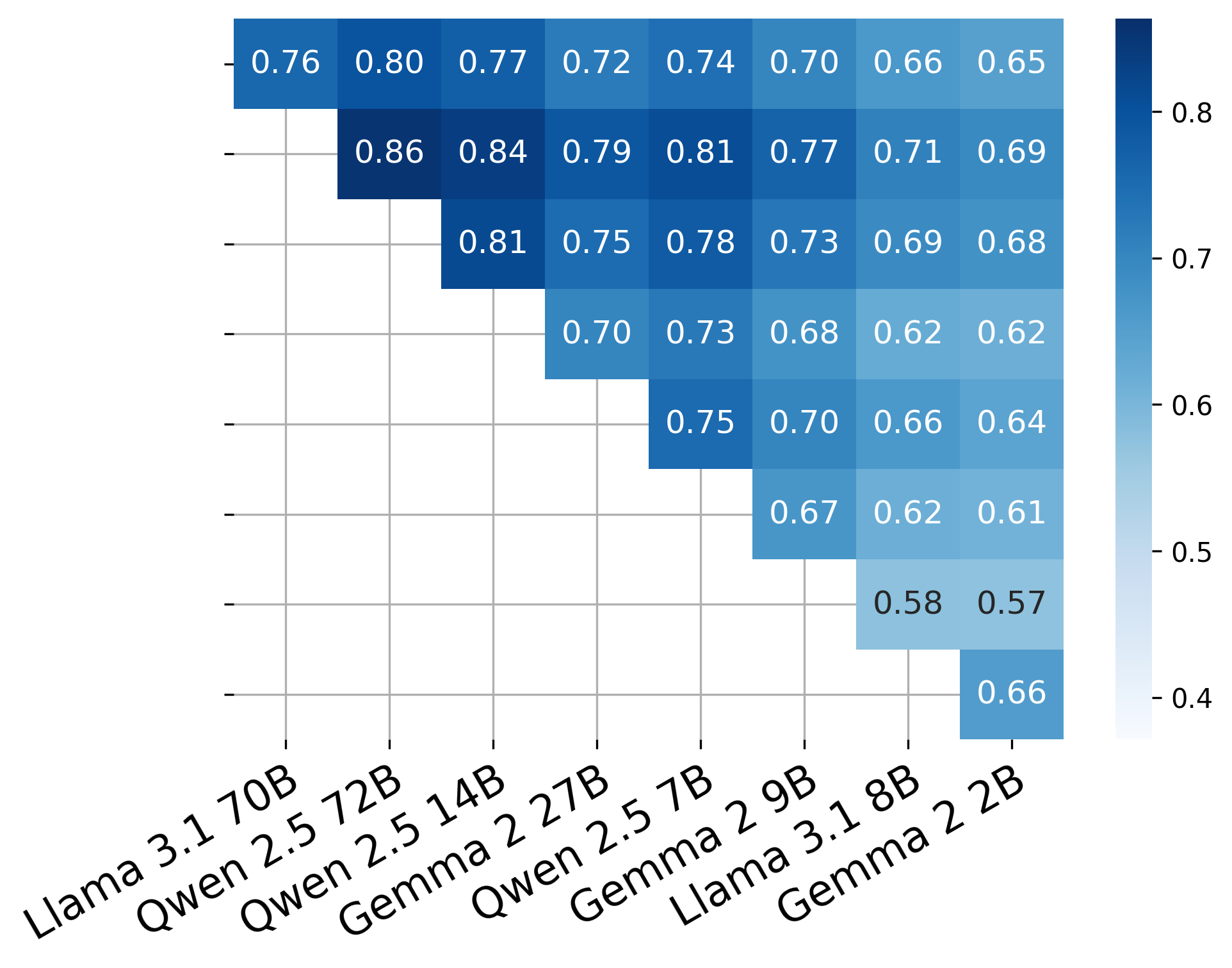}
    \caption{Qwen 2.5 72B}
\end{subfigure}
\begin{subfigure}{0.228\textwidth}
    \centering
    \includegraphics[width=\textwidth]{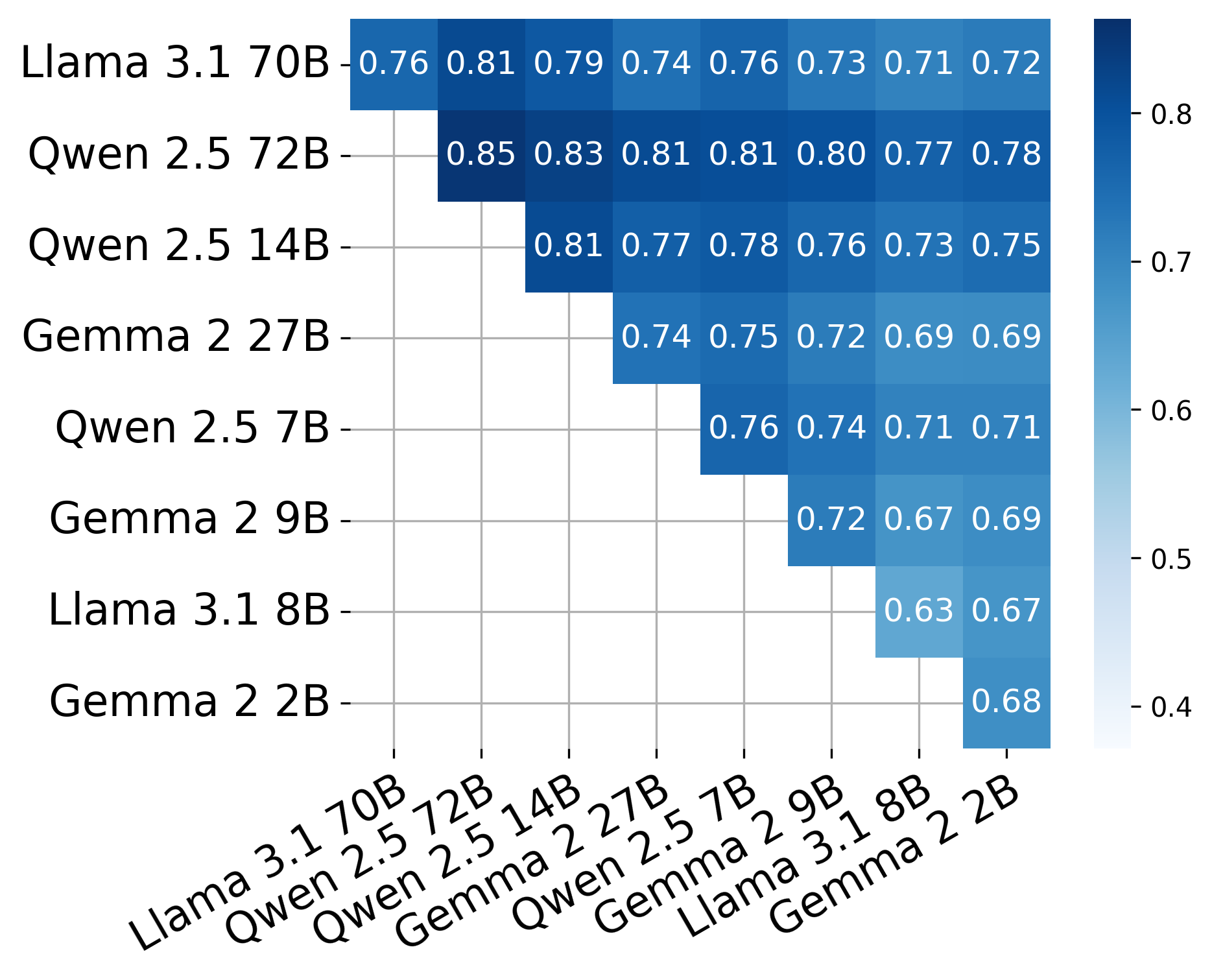}
    \caption{Qwen 2.5 14B}
\end{subfigure}
\begin{subfigure}{0.235\textwidth}
    \centering
    \includegraphics[width=\textwidth]{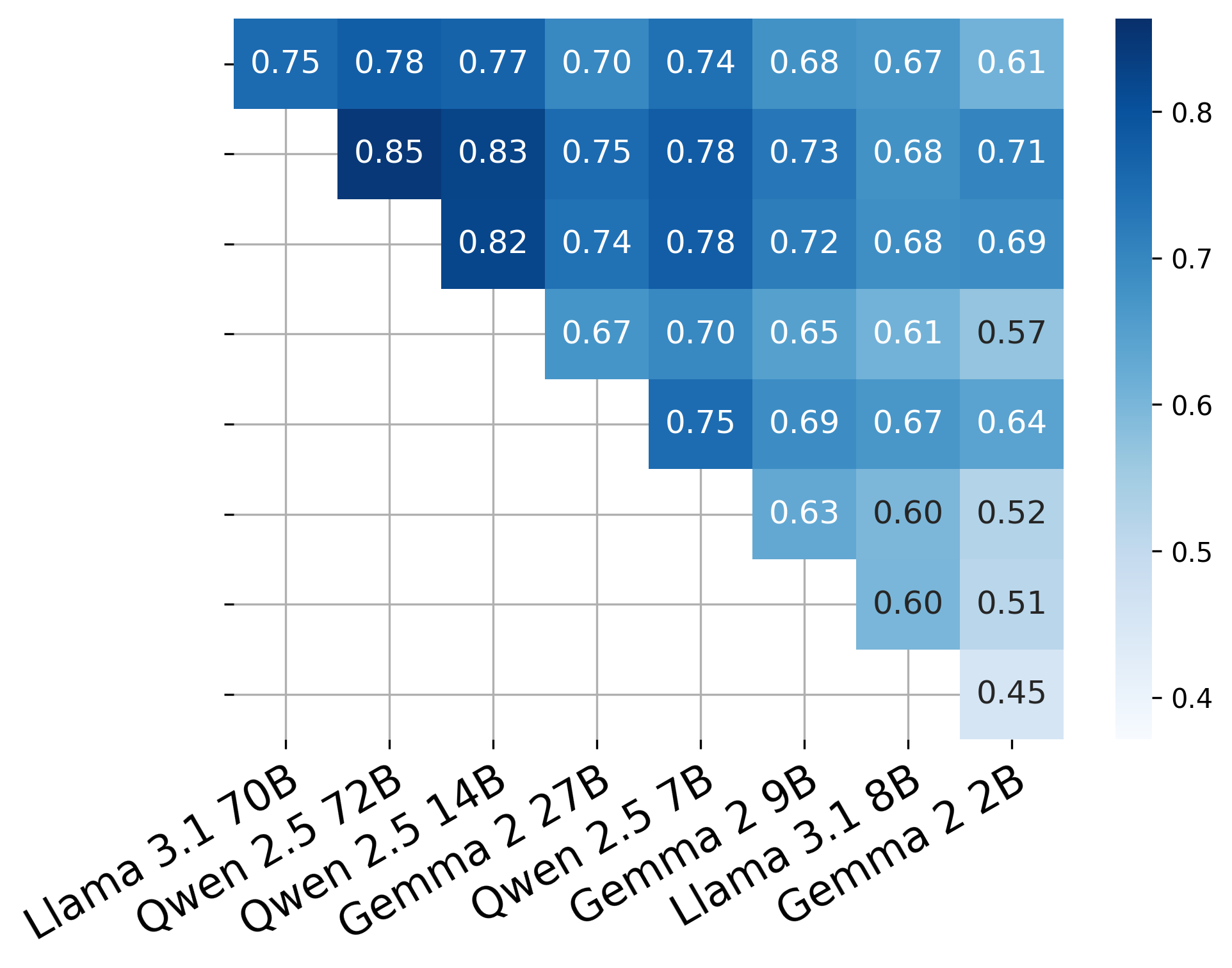}
    \caption{Gemma 2 27B}
\end{subfigure}
\caption{Judgment Performance $S^J_{A,B}$ of LLM judges on model pairs, averaged across datasets.}
\label{fig:performance_per_pair}
\end{figure}

\paragraph{Setup.} Figure \ref{fig:performance_per_pair} illustrates the judgement performance $S^J_{A,B}$ across model pairs $(A, B)$, indicating the probability of a correct judgement. The results are averaged over datasets and presented as an upper triangular matrix due to symmetry (we always present the answers in both possible orders and average performance).
We report the performance of all models used as judges in the Appendix \ref{app_sec:general_performance} in Table \ref{tab:performance_per_pair_full}.

\paragraph{Results.} 
The highest performance is achieved when two answers of Qwen 2.5 72B are compared which is the highest performing model (see task performance in Appendix \ref{app_subsec:task_performance})
In general, we observe that it is easier for the judge to make a correct judgment if candidate models are of higher quality.
This seems intuitive because such models likely structure and present their reasoning well, allowing a judge to compare solutions more easily.
Figure \ref{fig:confusion_matrices} gives an additional explanation. It shows that judges very reliably detect whether both answers are correct. When both models are capable, it is more likely that both give a correct answer, which makes it easier for LLM judges to classify correctly.

Interestingly, the judgment performances of Qwen 2.5 72B, Qwen 2.5 14B, and LLama 3.1 70B are very similar across pairs. The former possibly agree on a lot because of the similarity of training data and knowledge distillation \cite{knowledgedistillation}. The largest performance difference is that LLama 3.1 70B performs 10\% better when Qwen 2.5 72B and Gemma 2 2B are compared.

These results show that there is a relationship between the task performance of a candidate model and judgment performance. The following section will provide further analysis.


\section{Population-level Analysis: Judgements and Model Quality}
\label{sec:subset}

In this section, we investigate the relationship between LLM judgments and candidate LLM quality. First, we provide a statistical analysis where we use LLM task performance to explain the variance in LLM judgment performance.
Further, we focus on the subsets where the candidate models make exactly one correct and one incorrect prediction. We observe a strong statistical relationship between the difference in candidate task performances and judgment performances. 

\subsection{Can we explain Judgement Performance using Task Performance?}
\label{subsec:explain_judges}

A good indicator of the competence of a model on a specific dataset is its task performance. Clearly, there is a relationship between the quality of the involved models and the made judgments. We investigate the relationship between task performances (of candidate and judge models) and judgment performance.

\begin{table}
\centering
\resizebox{.49\textwidth}{!}{
\begin{tabular}{ccccc}
\toprule
 & Llama 3.1 70B & Qwen 2.5 72B & Qwen 2.5 14B & Gemma 2 27B \\
\midrule
$R^2$ & 0.89 & 0.87 & 0.85 & 0.93 \\
($p$-value) & (0.0) & (0.0) & (0.0) & (0.0) \\
\bottomrule
\end{tabular}
}
\caption{$R^2$ values for the regression models \textit{per judge} (first row) and corresponding $p$-values of the Overall F-Test (second row). All $R^2$ values are statistically significant on the 5\% level.}
\label{tab:exp_r2_overview}
\end{table}

\paragraph{Setup.} We fit multiple different linear regression models using the judgment performances as the target variables $Y$, including all variations of judges, model pairs $(A, B)$, and datasets $D$. Regarding the covariates $\mathbf{X}$ in the model, we solely use the task performances $S_X, X \in \{J, A, B\}$ of judge and candidate models, to predict judgment performance. 
Since we are not specifically interested in the individual features' effects, but rather in their ability to explain the variation of judgment performance, we rely on the coefficient of determination, $R^2$, for evaluation \citep[][ see Appendix \ref{a:stats}]{fahrmeir2013regression}.

\paragraph{Results.} The results are shown in Table \ref{tab:exp_r2_overview} (excluding data sets from the probability formulas for simplicity). We observe that the performance-related features of the models can explain the variation in the judgment performance (all $R^2$ values greater or equal than 0.85), very well. Logically, $S_A$ and $S_B$, have significant\footnote{We test statistical significance using an Overall-F-Test for each fitted model. Further details are in Appendix \ref{a:stats}.} explanatory power for judgment performance, as they encompass all correct answers. 












\subsection{Are LLM judges biased towards LLMs of higher quality?}
\label{sec:judge_bias}

To get a better understanding of whether there is a bias of LLM judges towards LLMs of higher quality, we investigate the subset where one candidate answer is correct and the other candidate answer is incorrect.
This subset is of the highest practical relevance.
The goal is to investigate the relationship between the task performances of the candidate models and the judge's performance.

\begin{figure}[t]  
\centering
\begin{subfigure}{0.246\textwidth}  
\centering
\includegraphics[width=\textwidth]{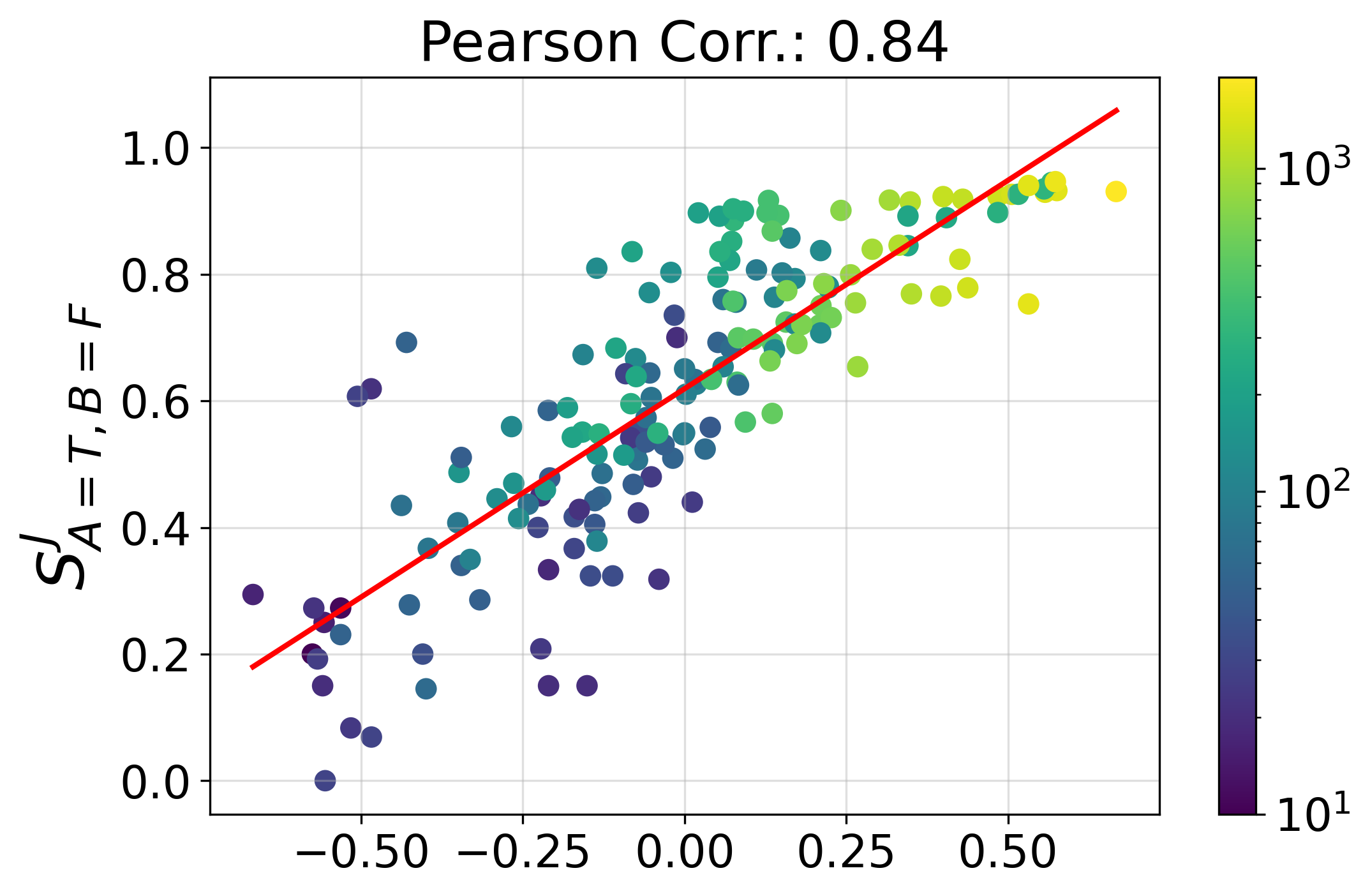}
\caption{Llama 3.1 70B}
\end{subfigure}
\begin{subfigure}{0.229\textwidth}  
\centering
\includegraphics[width=\textwidth]{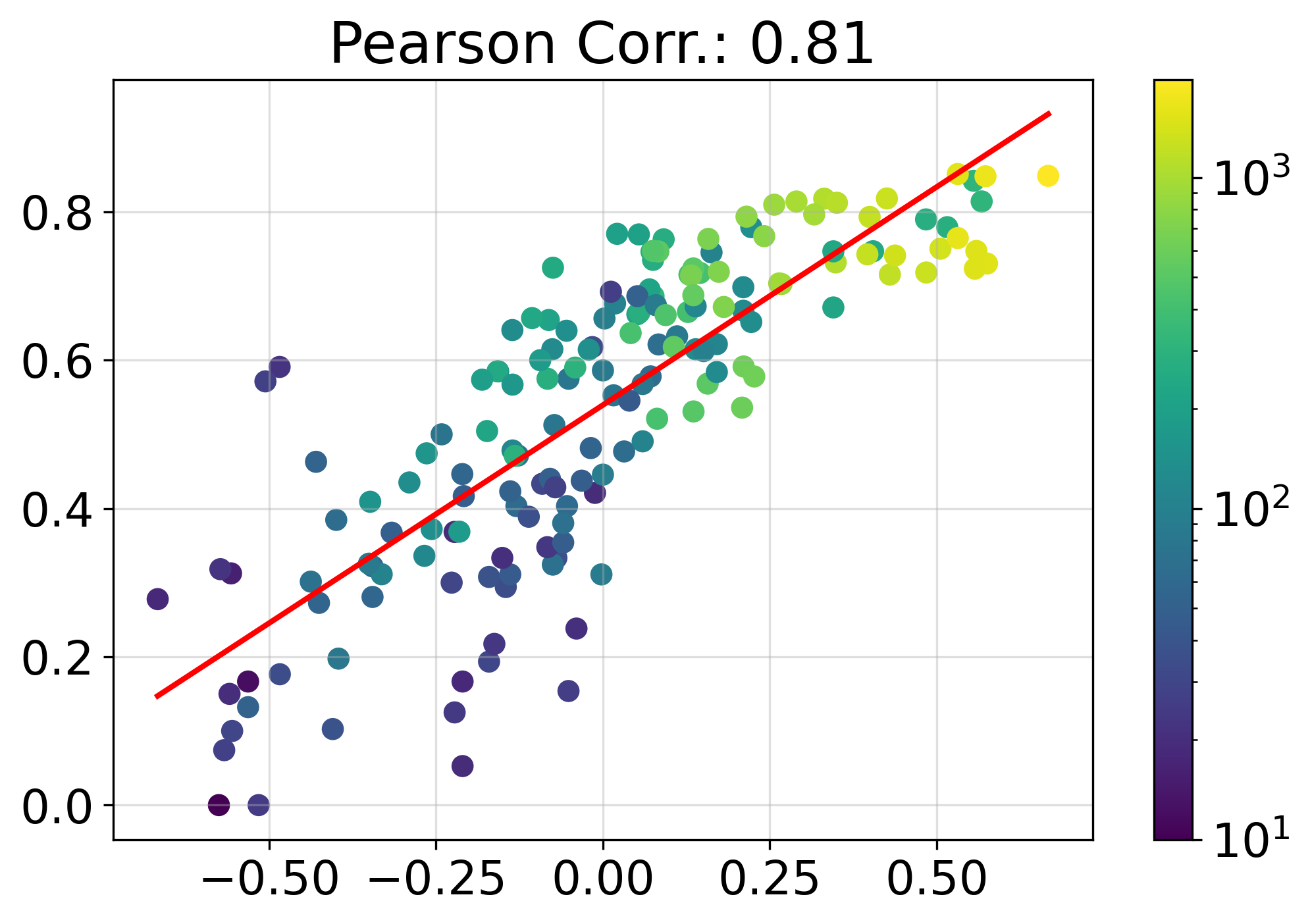}
\caption{Qwen2.5 72B}
\end{subfigure}
\begin{subfigure}{0.246\textwidth}
\centering
\includegraphics[width=\textwidth]{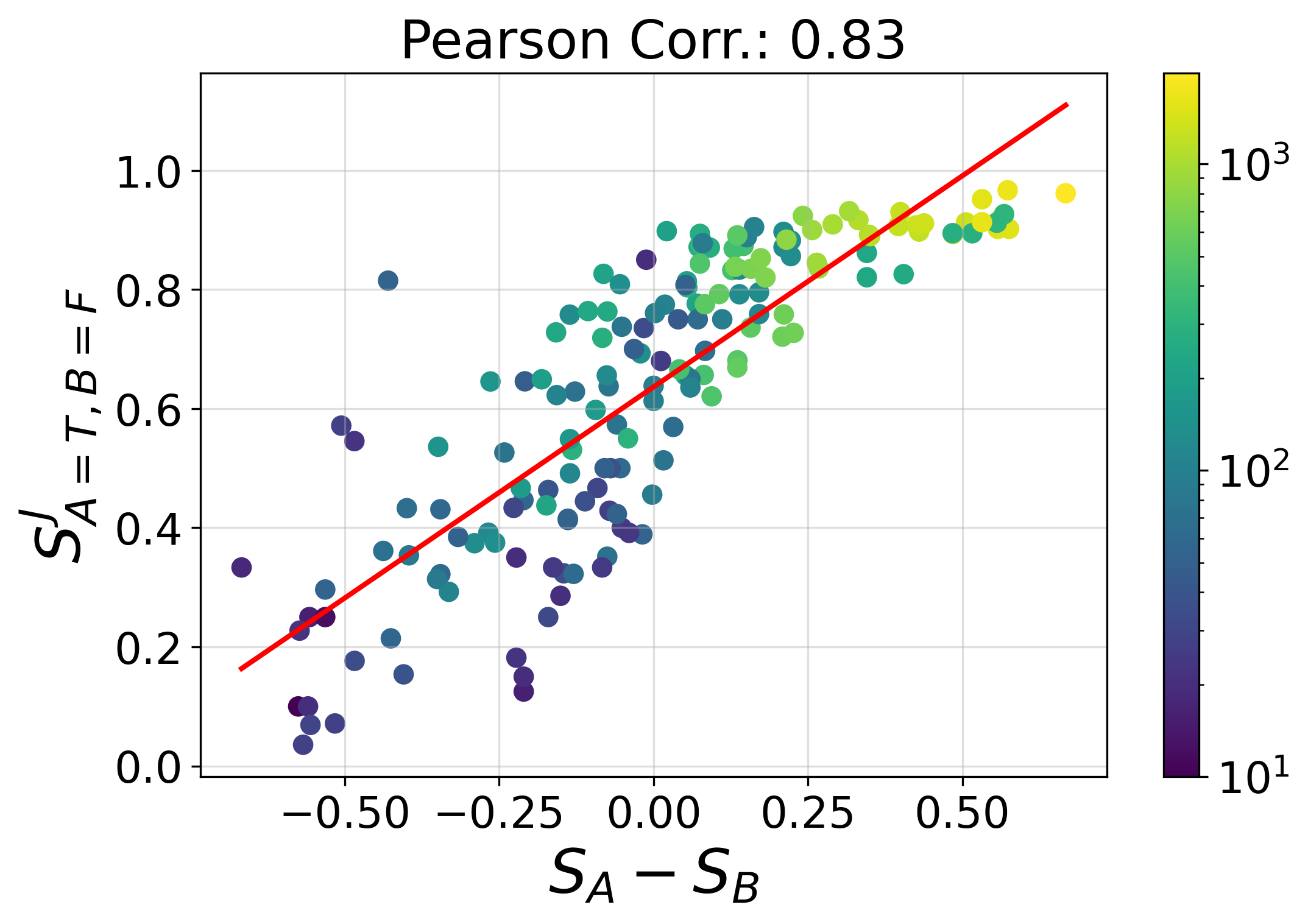}
\caption{Qwen 2.5 14B}
\end{subfigure}
\begin{subfigure}{0.229\textwidth}  
\centering
\includegraphics[width=\textwidth]{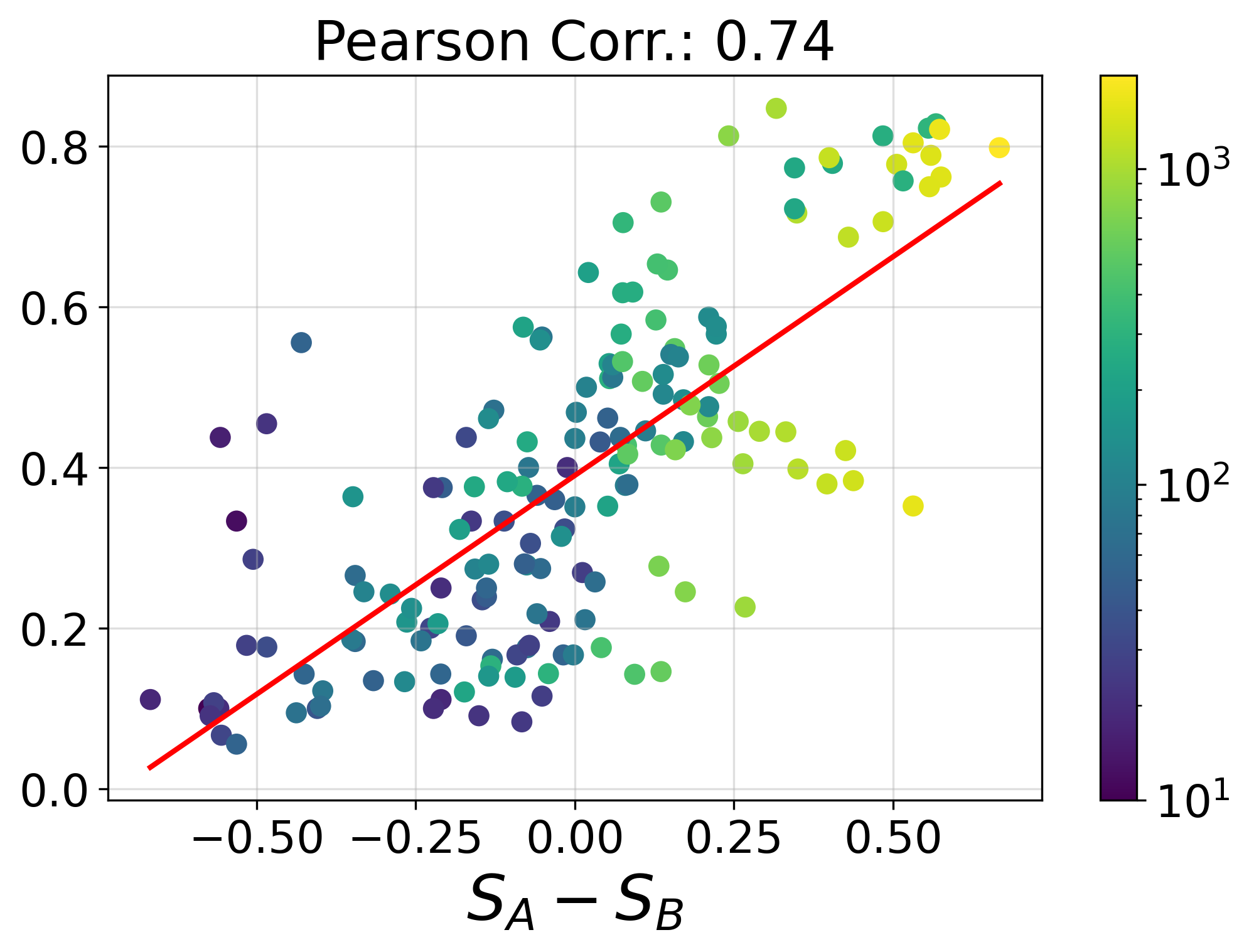}
\caption{Gemma 2 27B}
\end{subfigure}
\caption{Judges' accuracy vs. performance gap between two candidate models $A$ and $B$. Each point represents a subset where $A$ is correct, and $B$ is incorrect. The color reflects the size of these subsets.}
\label{fig:performance_diff_vs_accuracy}  
\end{figure}

\paragraph{Setup.} For all model pairs $(A, B), A \neq B$ we analyze subsets where $A$'s solutions are correct, and $B$'s solutions are incorrect, and call it 1-correct. Note that we can always order $A$ and $B$ this way. 
Each plot in Figure \ref{fig:performance_diff_vs_accuracy} shows the relationship between judge performance on the 1-correct subset (Y-axis) and \textit{candidate model performance gap} of $A$ and $B$, i.e., $S_A - S_B$ (X-axis). The color of the points indicate the size of the particular subset of samples.
Examples of these subsets and their corresponding performances are in Appendix \ref{app_sec:subset}.

\paragraph{Results.} The analysis reveals a strong correlation (Pearson's $r^2 > 0.78$) between judgement performance and candidate model performance gap. For the rest of this section, we call the model of higher performance on a dataset the \textit{more competent model}. I.e., if the performance gap is larger than $0$, the model giving the correct answer ($A$) is the more competent model. If the correct model is the more competent model, the judgment performance on the subset is higher, e.g., for LLama 3.1 70B, sometimes approaching 100\%. If the performance gap is more positive, it is easier to choose the correct answer. On the other hand, if the less competent model gives the correct answer, judgment performance is low, often lower than $20\%$. 

We infer that LLM judges are biased towards models of higher task performance. This finding aligns with previous research identifying self-bias \cite{xu2024prideprejudicellmamplifies, panickssery2024llmevaluatorsrecognizefavor, liu2024llmsnarcissisticevaluatorsego}, as judge LLMs are typically of higher quality than the judged models.
We hypothesize that this bias arises because more competent models articulate their responses more convincingly and exhibit a specific writing style, thereby misleading the judges.

However, models of higher task performance typically answer correctly more often (as indicated by the color of the points in Figure \ref{fig:performance_diff_vs_accuracy}.
\section{Sample-level analysis: Judgments and Stylistic Patterns}
\label{sec:prediction}

In Section \ref{sec:subset}, we found that the quality of a candidate LLM (as indicated by the task performance) correlates with the made judgment. We hypothesize that models of higher quality exhibit a particular style of expressing themselves and judges partially base their judgment on the incorporated textual cues. 
Motivated by recent work in machine-generated text detection which finds that LLMs often exhibit certain styles \cite{wu-aji-2025-style} or patterns \cite{shaib-etal-2024-detection}, we aim to gain a deeper understanding of whether shallow or even content-independent patterns affect the final judgment.

\paragraph{Setup.} We separate all judgments each judge made into training and test splits and train two classifiers. The test accuracy is reported in Table \ref{tab:linguistic_prediction}.
We use two types of features. First, we use TF-IDF embeddings. Secondly, we use N-Grams of part-of-speech (POS) tags, motivated by \citet{shaib-etal-2024-detection} who show and investigate the distinct occurrence of such in LLM-generated text. Given two candidate answers, we create two independent feature sets and concatenate those. Then a logistic regression and a RandomForest classifier \cite{breimanrandomforests} are trained on these concatenated features. Find more information in Appendix \ref{app_sec:sample_level}.

\paragraph{Results.} We observe that the models achieve a performance between approximately 70\% and 75\%. This indicates that structural information (POS tags) and word choice (TF-IDF) are important factors in understanding the patterns behind the behavior of LLM judges. The ground truth judgment distribution is shown in Appendix \ref{app_sec:sample_level}.

\begin{table}[t]
\centering
\resizebox{.49\textwidth}{!}{
\begin{tabular}{ll|cccc}
\toprule
Features Model &  & Llama 3.1 70B & Qwen 2.5 72B & Qwen 2.5 14B & Gemma 2 27B \\
\midrule
POS & LR & 72.79 & 69.66 & 72.33 & 70.19 \\
 & RF & 71.71 & 69.77 & 71.89 & 69.18 \\
TF-IDF & LR & 75.75 & 73.65 & 75.12 & 72.27 \\
  & RF & 75.65 & 71.05 & 75.79 & 70.58 \\
\bottomrule
\end{tabular}
}
\caption{Accuracy of predicting LLM judges' decisions using Logistic Regression (LR) and Random Forest (RF) classifiers based on N-Grams of either POS tags or TF-IDF features.}
\label{tab:linguistic_prediction}
\end{table}

Nevertheless, these results suggest that decision-making is a multi-faceted process. While specific shallow cues hold influence, a substantial portion of the decision-making process (25\%-30\%) can not be predicted this way and is based on other contextual factors which could include reasoning or noise.
\section{Usage recommendations}
\label{sec:usage_recs}

Lastly, we aim to give some usage recommendations. We start by analyzing two applied questions, namely, whether LLM judges can identify models of higher task performance and whether LLMs should be used to improve task performance. In the end, we discuss those results, connecting them to the overall insights of this paper.

\subsection{Do judges identify better models?}

An essential application of LLM judges is whether they can accurately identify which model performs better for a given task. This is crucial if we want to rank LLMs by their capabilities or if a practitioner wants to decide which model to deploy.

\begin{figure}
    \centering
    \includegraphics[width=0.97\linewidth]{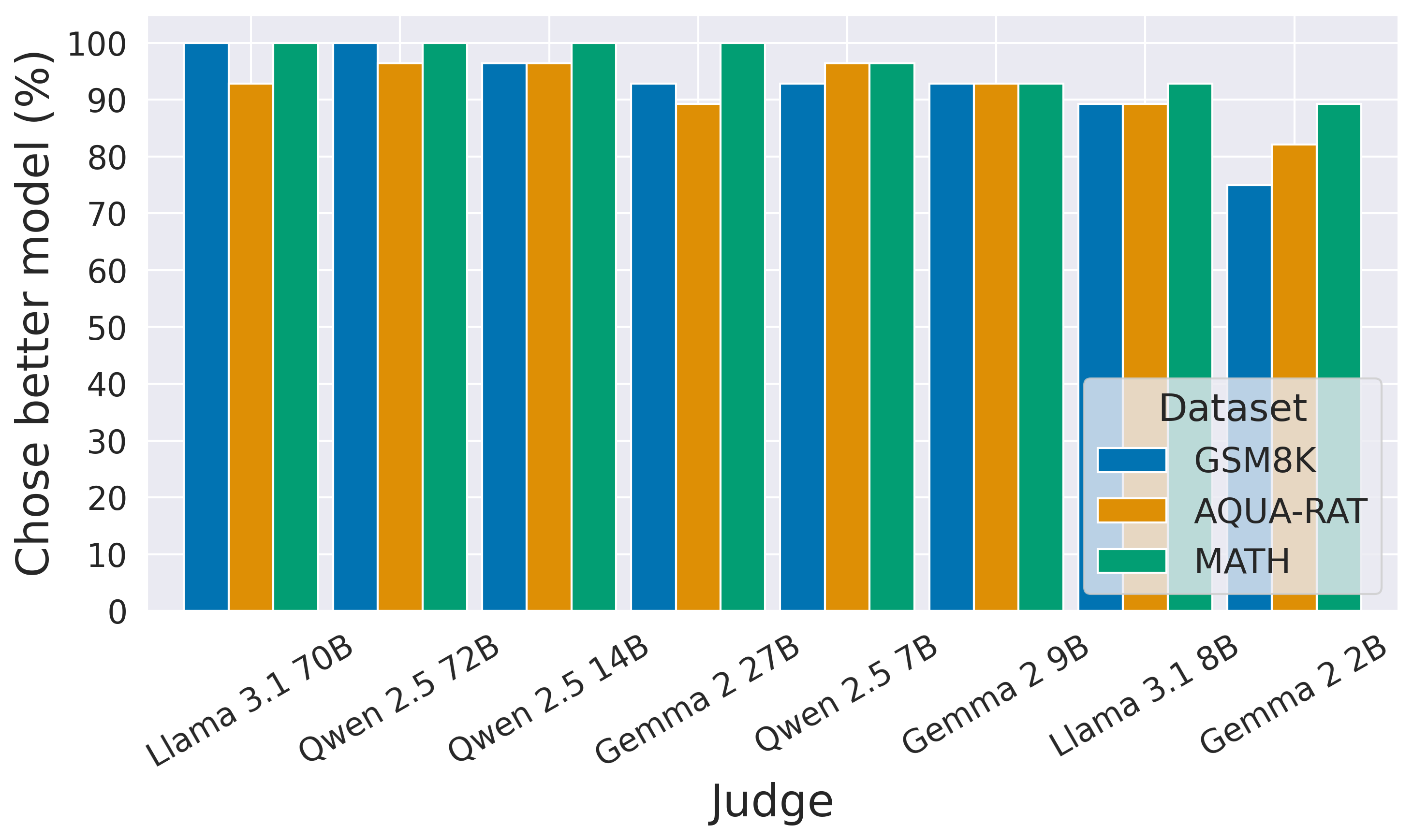}
    \caption{Percentage of model pairs $(A, B)$ where a judge picks a better model $A$ (meaning $S_A > S_B$), by selecting more answers of $A$ than from $B$.}
    \label{fig:chooses_better_model}
\end{figure}

\paragraph{Setup.} We evaluate which model a judge perceives as better by measuring the frequency of how often a judge selects the answer of a specific model. Formally, let $(A, B)$ be a candidate model pair where we assume that $A$ has higher task performance, i.e. $S_A > S_B$. If the judge chooses $A$ more often, we say a judge correctly determines $A$ to be better than $B$. For this analysis, we determine the proportion of model pairs $(A, B)$ for which the judge chooses $A$ over $B$ for all pairs $(A, B), S_A > S_B$ as shown in Figure \ref{fig:chooses_better_model}. 

\paragraph{Results.}
We observe that all tested large models consistently select the more competent model, i.e., the model with higher task performance. Also, the small models with $7$-$9$B parameters choose the correct model in over $90$\% of the cases. 
In general, it seems to be the hardest on the AQUA-RAT dataset. This is also the hardest dataset in Case (4), in Table \ref{tab:performance_per_dataset}, where exactly one answer is correct.
Note that the bias found in Section \ref{sec:judge_bias} is not necessarily problematic for this specific use case, because a bias towards the more competent model supports a correct outcome of this experiment. 


\subsection{Do judges elicit task improvement?}

Another interesting question of practical relevance is whether it makes sense to use LLM judges to improve task performance. One use case is the application of LLM judges in agentic systems where LLM judges might serve as a dedicated unit in a system.
Another use case is the subsequent usage of the answers chosen by the judge for self-training \cite{yuan2024selfrewardinglanguagemodels}.

\begin{figure}
\centering
\begin{subfigure}{0.223\textwidth}
    \centering
    \includegraphics[width=\textwidth]{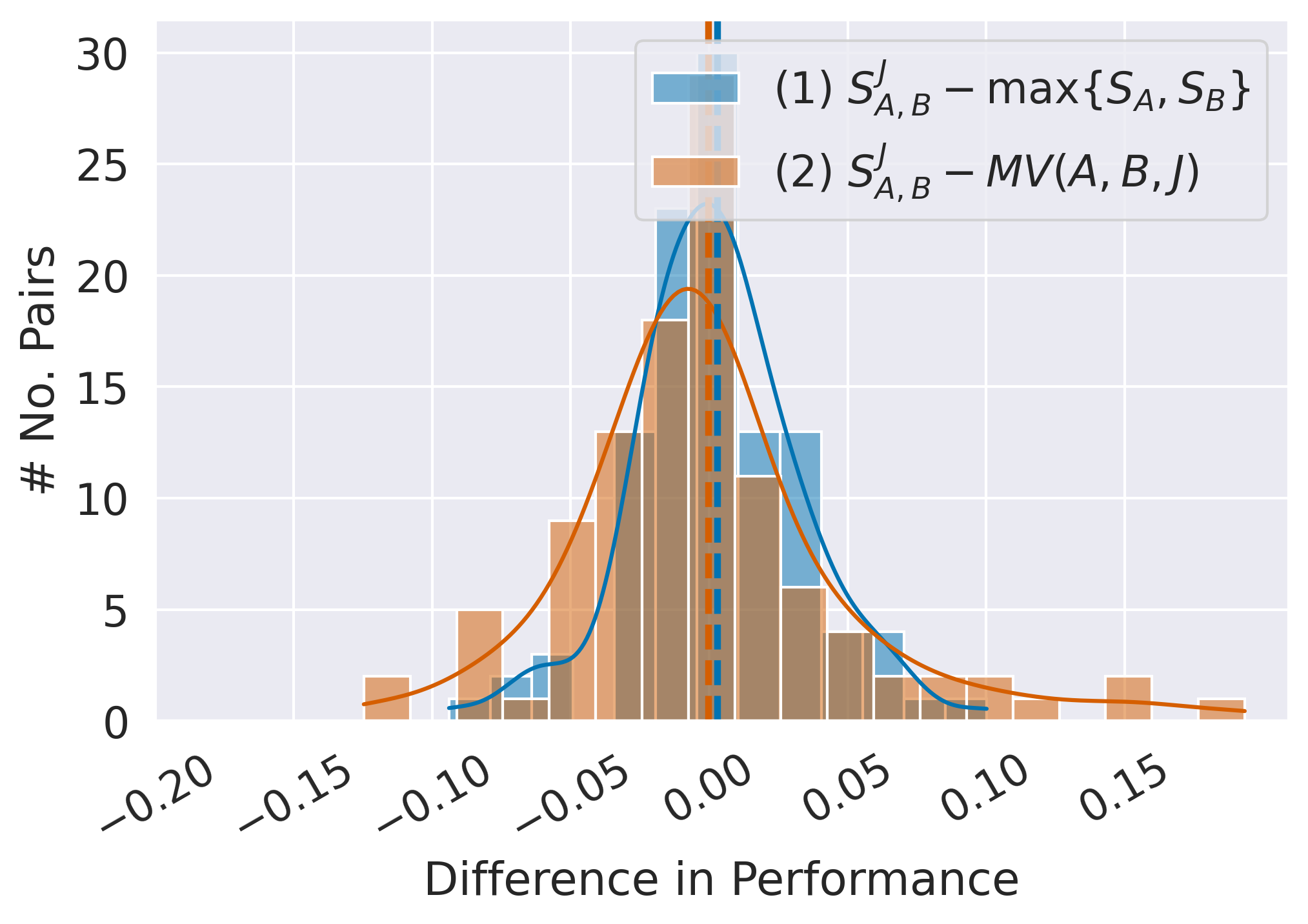}
    \caption{LLama3.1 70B Full}
\end{subfigure}
\begin{subfigure}{0.226\textwidth}
    \centering
    \includegraphics[width=\textwidth]{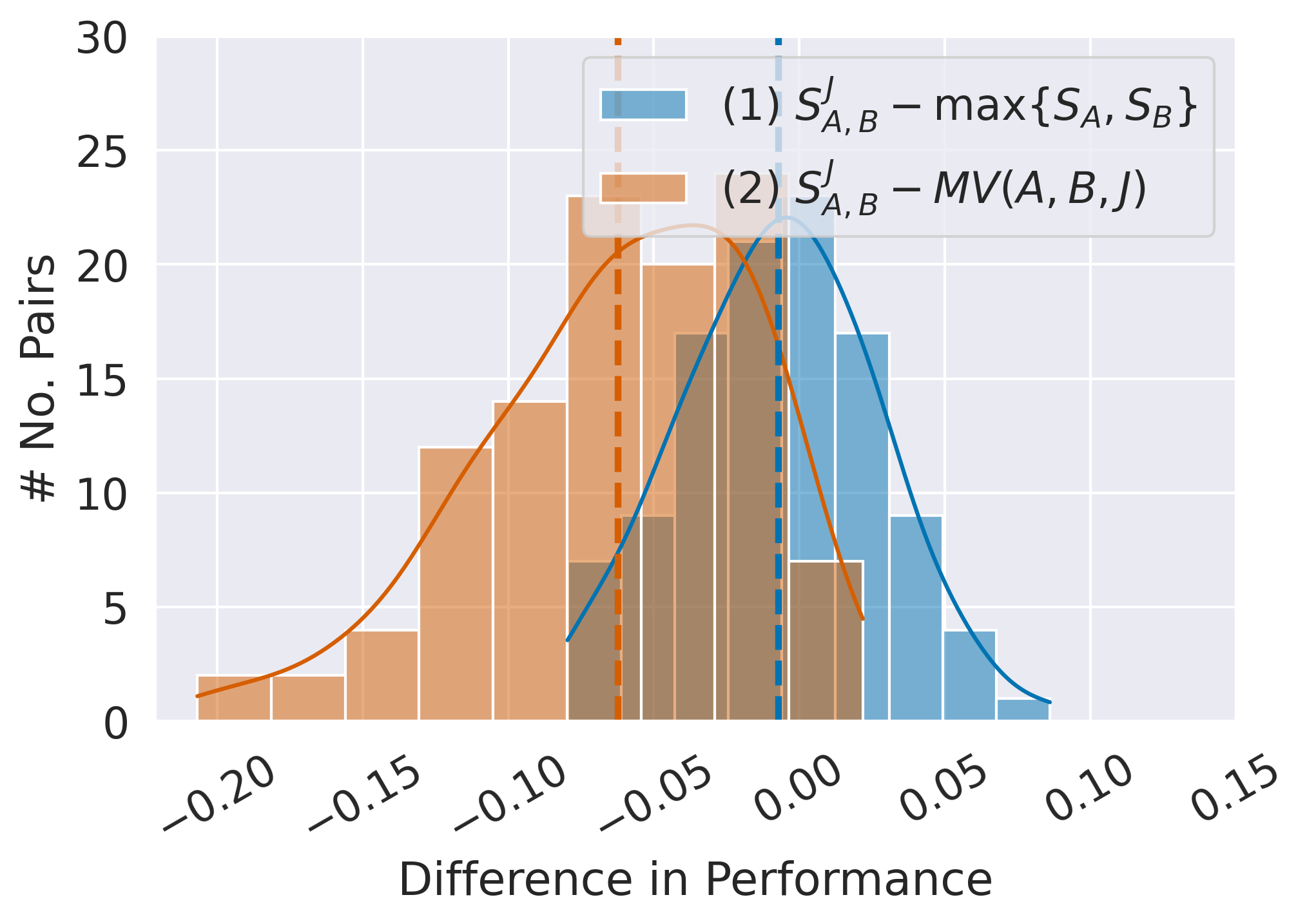}
    \caption{Qwen 2.5 72B Full}
\end{subfigure}
\begin{subfigure}{0.23\textwidth}
    \centering
    \includegraphics[width=\textwidth]{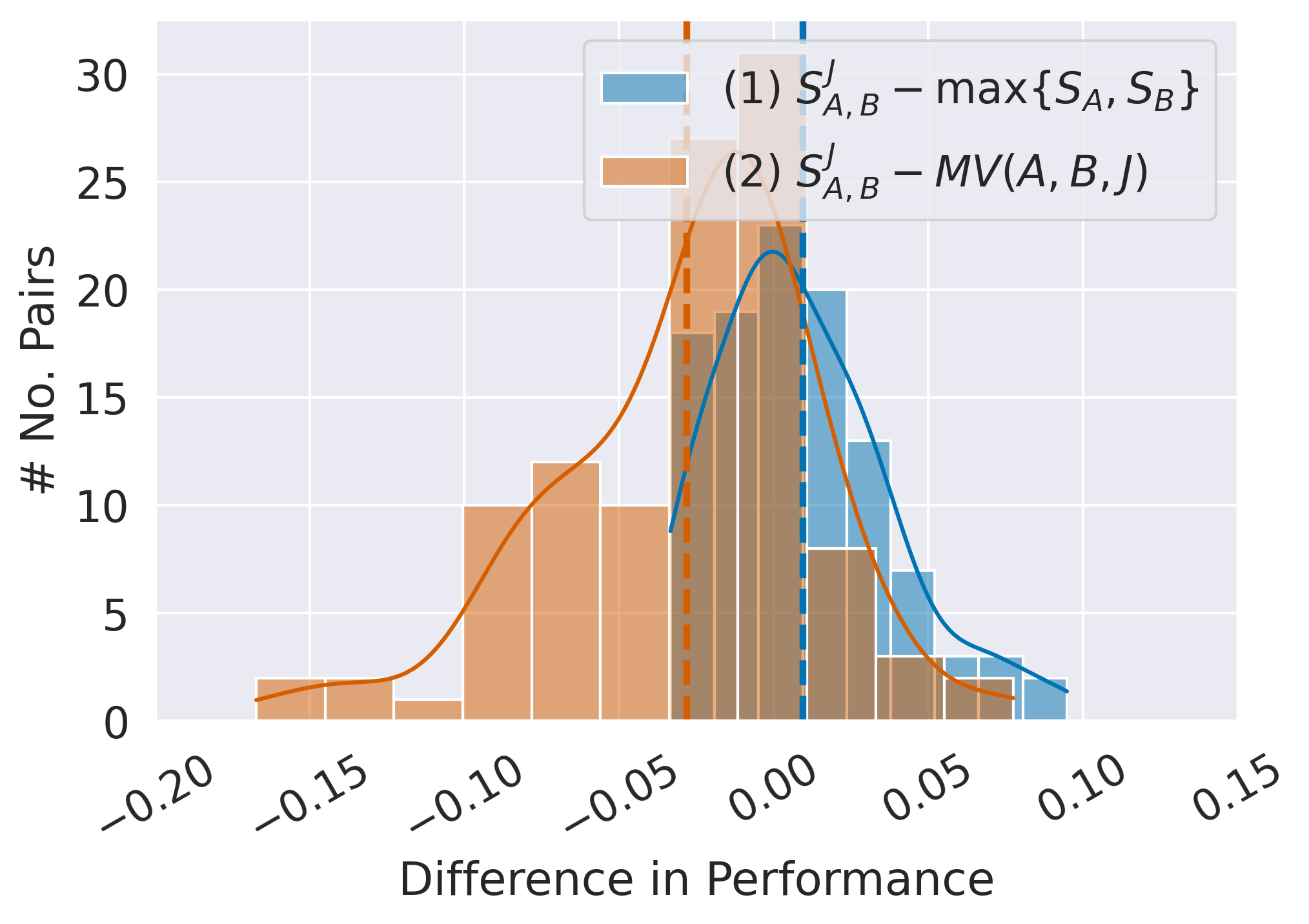}
    \caption{Qwen 2.5 14B Full}
\end{subfigure}
\begin{subfigure}{0.226\textwidth}
    \centering
    \includegraphics[width=\textwidth]{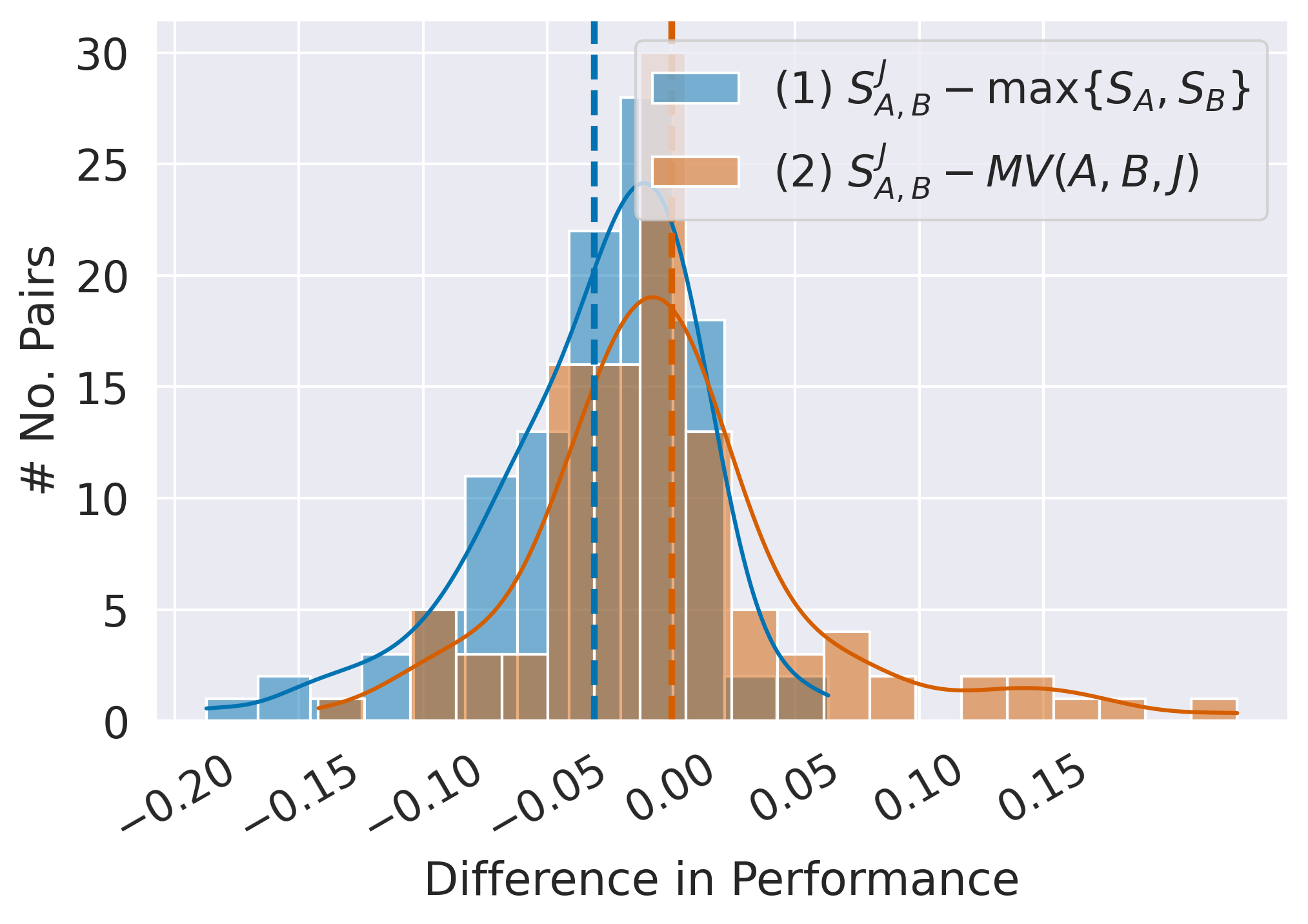}
    \caption{Gemma 2 27B}
\end{subfigure}
\caption{The Y-axis describes the number of model pairs $A, B$ where the answers chosen by the judge achieve a higher task performance than the performances of the individual models (blue) or than the majority vote (MV) of answers of $A, B$ and $J$ as candidate answer generator (red).
The X-axis describes the performance difference. A value of, e.g., $x=0.05$ means the answers chosen by the judge result in a 5\% (absolute) performance increase.}
\label{fig:improves_performance}
\end{figure}

\paragraph{Setup.} We separate the analysis into two questions. In Case (1), we evaluate whether the answers chosen by the judge result in a better performance than the individual models. Formally, for all pairs of models $(A, B)$, we plot the difference of performance of chosen answers, $C^J_{A,B}$ and maximal single candidate model performance $\max \{S_A, S_B \}$ in blue in a bar chart in Figure \ref{fig:improves_performance}.
Secondly, in Case (2), we test whether it makes more sense to use the judge model to generate a candidate answer $J$ and then take the majority vote across all three answers. Therefore we plot the performance difference of  $C^J_{A,B} - \text{MV}(A, B, J)$ in orange in a bar chart, where $\text{MV}(A, B, J)$ is the performance of the majority vote across all three answers.


\paragraph{Results.} 
In general, we observe that the performance differences are almost following a normal distribution. In Case (1), the distribution has a mean value (dashed line) slightly larger than 0 for LLama 3.1 70B (0.3) and Qwen 2.5 14B (0.9). That means that, on average, the answers chosen by the judge result in slightly increased performance, e.g., an increase from 40\% accuracy to 40.9\% accuracy. In Case (2), the mean value is never larger than 0, meaning that the majority vote is more likely to be better than the answer chosen by the judge. Especially for Qwen 2.5 14B and Qwen 2 72B, it is more viable to use the majority voting strategy.

\subsection{Discussion}

Our analysis of LLM judges on mathematical reasoning tasks reveals several insights for practitioners, which we discuss in the following. We separate our discussion into the sample level, i.e., the interpretation of a single prediction, and aggregate level, i.e., the interpretation of a set of predictions.

\paragraph{Sample level. } In Table \ref{tab:performance_per_dataset}, we find that LLM judges often achieve a strong judgment performance ($S^J_{A,B}$ $>80\%$ accuracy) across tasks. While this is a solid classification performance, it means that the prediction is wrong in 20\% of the cases which limits practical applicability. 
In Section \ref{sec:general_performance}, we observe that LLM judges demonstrate high precision when identifying correct answers from both models. This might be valuable for filtering samples and curating training data or, e.g., self-training. Nevertheless, one has to be careful how to use these because correctly judged samples are biased towards simple samples.
In summary, we do not recommend fully relying on individual LLM judgments, especially not in high-stakes domains such as legal or health care.

\paragraph{Aggregate level. } 
As shown in Figure \ref{fig:chooses_better_model}, we find that LLM judges are consistently able to select or rank models by their task performance.
This is supported by Section \ref{subsec:explain_judges} where we show that a simple linear model can explain a high share of the variance in judgment performance, given individual task performances, suggesting that the performance difference of two candidate models is linearly linked to the judgment outcome. 

In summary, our results suggest that LLM judges are more effective and consistent at aggregate-level comparisons than instance-level judgments, for example when ranking or selecting which LLM is better for a particular task when no ground truth data is available. 

\section{Conclusion}

We conduct a thorough analysis of LLM judges on mathematical reasoning tasks. We evaluate the judgment performance of eight models of different sizes on three datasets. 
We find that larger judge models generally outperform smaller judge models and that judges can reliably detect whether both answers are correct.
Our analysis reveals a strong correlation between judgments and task performance, indicating that judges tend to choose models of higher quality even if their answers are incorrect.
We hypothesize that LLM judges partially base their decisions on linguistic cues in contrast to the reasoning within the answers. 
We support this hypothesis with our experiments showing that 70\% of the judges' decisions can be predicted using simple linguistic features such as N-grams of part-of-speech tags.
Lastly, our analysis finds that LLM judges reliably detect LLMs of higher task performance but are not reliably useable to improve task performance.
Our results show that LLM judges contain biases and suggest that practitioners should not blindly trust LLM judges. We advise practitioners to carefully decide whether LLM judges should be used in their particular application.

With this work, we set the stage for further research to investigate how to understand, use, and improve LLM judges.

\newpage
\section{Limitations}
 
Our analysis is primarily focused on mathematical reasoning datasets, which allows us to explore judgments through the lens of verifiability, i.e., problems that have a definitely correct answer. While this approach provides valuable insights, it limits the generalizability of our findings to other tasks or domains. 
Nevertheless, we want to emphasize the importance of the class of verifiable tasks. For instance, there is currently a focus on training so-called large reasoning models, which demonstrate significant progress in solving complex problems such as coding or maths. It is a possibility that an increased capability of LLMs on verifiable tasks fuels scientific progress.

In our experiments, we focus on testing a single, specific prompt. It is common knowledge that LLMs are highly sensitive to variations in prompt phrasing, which can substantially influence their performance. However, the resources available to us do not allow us to meet the computational demands necessary to run our experiments with multiple prompts. 
Further, our impression is that it is a custom approach to conduct LLM studies using single prompts, as they are typically indicative of behavior. Therefore we decided to run our analysis on full datasets with a single prompt instead of using subsets of datasets with variations of the prompt with mostly the same content.


\bibliography{custom}

\begin{thebibliography}{42}
\expandafter\ifx\csname natexlab\endcsname\relax\def\natexlab#1{#1}\fi

\bibitem[{AI@Meta(2024)}]{llama3modelcard}
AI@Meta. 2024.
\newblock \href {https://github.com/meta-llama/llama3/blob/main/MODEL_CARD.md}
  {Llama 3 model card}.

\bibitem[{Bai et~al.(2022)Bai, Kadavath, Kundu, Askell, Kernion, Jones, Chen,
  Goldie, Mirhoseini, McKinnon et~al.}]{bai2022constitutional}
Yuntao Bai, Saurav Kadavath, Sandipan Kundu, Amanda Askell, Jackson Kernion,
  Andy Jones, Anna Chen, Anna Goldie, Azalia Mirhoseini, Cameron McKinnon,
  et~al. 2022.
\newblock Constitutional ai: Harmlessness from ai feedback.
\newblock \emph{arXiv preprint arXiv:2212.08073}.

\bibitem[{Bangalore Principles, 2002()}]{bangalore_principles}
Bangalore Principles, 2002. 2002.
\newblock \href {https://www.judicialintegritygroup.org/jig-principles} {The
  bangalore principles of judicial conduct}.
\newblock Available from the Judicial Integrity Group website.

\bibitem[{Bavaresco et~al.(2024)Bavaresco, Bernardi, Bertolazzi, Elliott,
  Fernández, Gatt, Ghaleb, Giulianelli, Hanna, Koller, Martins, Mondorf,
  Neplenbroek, Pezzelle, Plank, Schlangen, Suglia, Surikuchi, Takmaz, and
  Testoni}]{bavaresco2024llmsinsteadhumanjudges}
Anna Bavaresco, Raffaella Bernardi, Leonardo Bertolazzi, Desmond Elliott,
  Raquel Fernández, Albert Gatt, Esam Ghaleb, Mario Giulianelli, Michael
  Hanna, Alexander Koller, André F.~T. Martins, Philipp Mondorf, Vera
  Neplenbroek, Sandro Pezzelle, Barbara Plank, David Schlangen, Alessandro
  Suglia, Aditya~K Surikuchi, Ece Takmaz, and Alberto Testoni. 2024.
\newblock \href {http://arxiv.org/abs/2406.18403} {Llms instead of human
  judges? a large scale empirical study across 20 nlp evaluation tasks}.

\bibitem[{Breiman(2001)}]{breimanrandomforests}
Leo Breiman. 2001.
\newblock \href {https://doi.org/10.1023/A:1010933404324} {Random forests}.
\newblock \emph{Mach. Learn.}, 45(1):5–32.

\bibitem[{Cobbe et~al.(2021)Cobbe, Kosaraju, Bavarian, Chen, Jun, Kaiser,
  Plappert, Tworek, Hilton, Nakano, Hesse, and Schulman}]{cobbe2021gsm8k}
Karl Cobbe, Vineet Kosaraju, Mohammad Bavarian, Mark Chen, Heewoo Jun, Lukasz
  Kaiser, Matthias Plappert, Jerry Tworek, Jacob Hilton, Reiichiro Nakano,
  Christopher Hesse, and John Schulman. 2021.
\newblock Training verifiers to solve math word problems.
\newblock \emph{arXiv preprint arXiv:2110.14168}.

\bibitem[{Doddapaneni et~al.(2024)Doddapaneni, Khan, Verma, and
  Khapra}]{doddapaneni-etal-2024-finding}
Sumanth Doddapaneni, Mohammed Safi Ur~Rahman Khan, Sshubam Verma, and Mitesh~M
  Khapra. 2024.
\newblock \href {https://doi.org/10.18653/v1/2024.emnlp-main.911} {Finding
  blind spots in evaluator {LLM}s with interpretable checklists}.
\newblock In \emph{Proceedings of the 2024 Conference on Empirical Methods in
  Natural Language Processing}, pages 16279--16309, Miami, Florida, USA.
  Association for Computational Linguistics.

\bibitem[{Fahrmeir et~al.(2013)Fahrmeir, Kneib, Lang, Marx, Fahrmeir, Kneib,
  Lang, and Marx}]{fahrmeir2013regression}
Ludwig Fahrmeir, Thomas Kneib, Stefan Lang, Brian Marx, Ludwig Fahrmeir, Thomas
  Kneib, Stefan Lang, and Brian Marx. 2013.
\newblock \emph{Regression models}.
\newblock Springer.

\bibitem[{Gemma~Team et~al.(2024)Gemma~Team, Mesnard, Hardin, Dadashi,
  Bhupatiraju, Pathak, Sifre, Rivière, Kale, Love, Tafti, Hussenot, Sessa,
  Chowdhery, Roberts, Barua, Botev, Castro-Ros, Slone, Héliou, Tacchetti,
  Bulanova, Paterson, Tsai, Shahriari, Lan, Choquette-Choo, Crepy, Cer,
  Ippolito, Reid, Buchatskaya, Ni, Noland, Yan, Tucker, Muraru,
  Rozhdestvenskiy, Michalewski, Tenney, Grishchenko, Austin, Keeling,
  Labanowski, Lespiau, Stanway, Brennan, Chen, Ferret, Chiu, Mao-Jones, Lee,
  Yu, Millican, Sjoesund, Lee, Dixon, Reid, Mikuła, Wirth, Sharman, Chinaev,
  Thain, Bachem, Chang, Wahltinez, Bailey, Michel, Yotov, Chaabouni, Comanescu,
  Jana, Anil, McIlroy, Liu, Mullins, Smith, Borgeaud, Girgin, Douglas, Pandya,
  Shakeri, De, Klimenko, Hennigan, Feinberg, Stokowiec, hui Chen, Ahmed, Gong,
  Warkentin, Peran, Giang, Farabet, Vinyals, Dean, Kavukcuoglu, Hassabis,
  Ghahramani, Eck, Barral, Pereira, Collins, Joulin, Fiedel, Senter, Andreev,
  and Kenealy}]{gemmateam2024gemmaopenmodelsbased}
Google Gemma~Team, Thomas Mesnard, Cassidy Hardin, Robert Dadashi, Surya
  Bhupatiraju, Shreya Pathak, Laurent Sifre, Morgane Rivière, Mihir~Sanjay
  Kale, Juliette Love, Pouya Tafti, Léonard Hussenot, Pier~Giuseppe Sessa,
  Aakanksha Chowdhery, Adam Roberts, Aditya Barua, Alex Botev, Alex Castro-Ros,
  Ambrose Slone, Amélie Héliou, Andrea Tacchetti, Anna Bulanova, Antonia
  Paterson, Beth Tsai, Bobak Shahriari, Charline~Le Lan, Christopher~A.
  Choquette-Choo, Clément Crepy, Daniel Cer, Daphne Ippolito, David Reid,
  Elena Buchatskaya, Eric Ni, Eric Noland, Geng Yan, George Tucker,
  George-Christian Muraru, Grigory Rozhdestvenskiy, Henryk Michalewski, Ian
  Tenney, Ivan Grishchenko, Jacob Austin, James Keeling, Jane Labanowski,
  Jean-Baptiste Lespiau, Jeff Stanway, Jenny Brennan, Jeremy Chen, Johan
  Ferret, Justin Chiu, Justin Mao-Jones, Katherine Lee, Kathy Yu, Katie
  Millican, Lars~Lowe Sjoesund, Lisa Lee, Lucas Dixon, Machel Reid, Maciej
  Mikuła, Mateo Wirth, Michael Sharman, Nikolai Chinaev, Nithum Thain, Olivier
  Bachem, Oscar Chang, Oscar Wahltinez, Paige Bailey, Paul Michel, Petko Yotov,
  Rahma Chaabouni, Ramona Comanescu, Reena Jana, Rohan Anil, Ross McIlroy,
  Ruibo Liu, Ryan Mullins, Samuel~L Smith, Sebastian Borgeaud, Sertan Girgin,
  Sholto Douglas, Shree Pandya, Siamak Shakeri, Soham De, Ted Klimenko, Tom
  Hennigan, Vlad Feinberg, Wojciech Stokowiec, Yu~hui Chen, Zafarali Ahmed,
  Zhitao Gong, Tris Warkentin, Ludovic Peran, Minh Giang, Clément Farabet,
  Oriol Vinyals, Jeff Dean, Koray Kavukcuoglu, Demis Hassabis, Zoubin
  Ghahramani, Douglas Eck, Joelle Barral, Fernando Pereira, Eli Collins, Armand
  Joulin, Noah Fiedel, Evan Senter, Alek Andreev, and Kathleen Kenealy. 2024.
\newblock \href {http://arxiv.org/abs/2403.08295} {Gemma: Open models based on
  gemini research and technology}.

\bibitem[{Hendrycks et~al.(2021)Hendrycks, Burns, Kadavath, Arora, Basart,
  Tang, Song, and Steinhardt}]{hendrycksmath2021}
Dan Hendrycks, Collin Burns, Saurav Kadavath, Akul Arora, Steven Basart, Eric
  Tang, Dawn Song, and Jacob Steinhardt. 2021.
\newblock Measuring mathematical problem solving with the math dataset.
\newblock \emph{NeurIPS}.

\bibitem[{Hinton et~al.(2014)Hinton, Dean, and Vinyals}]{knowledgedistillation}
Geoffrey Hinton, Jeff Dean, and Oriol Vinyals. 2014.
\newblock Distilling the knowledge in a neural network.
\newblock In \emph{NIPS 2014 Deep Learning Workshop}, pages 1--9.

\bibitem[{Jiang et~al.(2023)Jiang, Sablayrolles, Mensch, Bamford, Chaplot,
  de~las Casas, Bressand, Lengyel, Lample, Saulnier, Lavaud, Lachaux, Stock,
  Scao, Lavril, Wang, Lacroix, and Sayed}]{jiang2023mistral7b}
Albert~Q. Jiang, Alexandre Sablayrolles, Arthur Mensch, Chris Bamford,
  Devendra~Singh Chaplot, Diego de~las Casas, Florian Bressand, Gianna Lengyel,
  Guillaume Lample, Lucile Saulnier, Lélio~Renard Lavaud, Marie-Anne Lachaux,
  Pierre Stock, Teven~Le Scao, Thibaut Lavril, Thomas Wang, Timothée Lacroix,
  and William~El Sayed. 2023.
\newblock \href {http://arxiv.org/abs/2310.06825} {Mistral 7b}.

\bibitem[{Jiang et~al.(2024)Jiang, Sablayrolles, Roux, Mensch, Savary, Bamford,
  Chaplot, de~las Casas, Hanna, Bressand, Lengyel, Bour, Lample, Lavaud,
  Saulnier, Lachaux, Stock, Subramanian, Yang, Antoniak, Scao, Gervet, Lavril,
  Wang, Lacroix, and Sayed}]{jiang2024mixtralexperts}
Albert~Q. Jiang, Alexandre Sablayrolles, Antoine Roux, Arthur Mensch, Blanche
  Savary, Chris Bamford, Devendra~Singh Chaplot, Diego de~las Casas, Emma~Bou
  Hanna, Florian Bressand, Gianna Lengyel, Guillaume Bour, Guillaume Lample,
  Lélio~Renard Lavaud, Lucile Saulnier, Marie-Anne Lachaux, Pierre Stock,
  Sandeep Subramanian, Sophia Yang, Szymon Antoniak, Teven~Le Scao, Théophile
  Gervet, Thibaut Lavril, Thomas Wang, Timothée Lacroix, and William~El Sayed.
  2024.
\newblock \href {http://arxiv.org/abs/2401.04088} {Mixtral of experts}.

\bibitem[{Kim et~al.(2024{\natexlab{a}})Kim, Shin, Cho, Jang, Longpre, Lee,
  Yun, Shin, Kim, Thorne, and Seo}]{kim2024prometheus}
Seungone Kim, Jamin Shin, Yejin Cho, Joel Jang, Shayne Longpre, Hwaran Lee,
  Sangdoo Yun, Seongjin Shin, Sungdong Kim, James Thorne, and Minjoon Seo.
  2024{\natexlab{a}}.
\newblock \href {https://openreview.net/forum?id=8euJaTveKw} {Prometheus:
  Inducing fine-grained evaluation capability in language models}.
\newblock In \emph{The Twelfth International Conference on Learning
  Representations}.

\bibitem[{Kim et~al.(2024{\natexlab{b}})Kim, Suk, Cho, Longpre, Kim, Yoon, Son,
  Cho, Shafayat, Baek, Park, Hwang, Jo, Cho, Shin, Lee, Oh, Lee, Ho, Joo, Ko,
  Lee, Chae, Shin, Jang, Ye, Lin, Welleck, Neubig, Lee, Lee, and
  Seo}]{kim2024biggen}
Seungone Kim, Juyoung Suk, Ji~Yong Cho, Shayne Longpre, Chaeeun Kim, Dongkeun
  Yoon, Guijin Son, Yejin Cho, Sheikh Shafayat, Jinheon Baek, Sue~Hyun Park,
  Hyeonbin Hwang, Jinkyung Jo, Hyowon Cho, Haebin Shin, Seongyun Lee, Hanseok
  Oh, Noah Lee, Namgyu Ho, Se~June Joo, Miyoung Ko, Yoonjoo Lee, Hyungjoo Chae,
  Jamin Shin, Joel Jang, Seonghyeon Ye, Bill~Yuchen Lin, Sean Welleck, Graham
  Neubig, Moontae Lee, Kyungjae Lee, and Minjoon Seo. 2024{\natexlab{b}}.
\newblock \href {http://arxiv.org/abs/2406.05761} {The biggen bench: A
  principled benchmark for fine-grained evaluation of language models with
  language models}.

\bibitem[{Koo et~al.(2023)Koo, Lee, Raheja, Park, Kim, and
  Kang}]{koo2023benchmarkingcognitivebiaseslarge}
Ryan Koo, Minhwa Lee, Vipul Raheja, Jong~Inn Park, Zae~Myung Kim, and Dongyeop
  Kang. 2023.
\newblock \href {http://arxiv.org/abs/2309.17012} {Benchmarking cognitive
  biases in large language models as evaluators}.

\bibitem[{Kwon et~al.(2023)Kwon, Li, Zhuang, Sheng, Zheng, Yu, Gonzalez, Zhang,
  and Stoica}]{kwon2023efficient}
Woosuk Kwon, Zhuohan Li, Siyuan Zhuang, Ying Sheng, Lianmin Zheng, Cody~Hao Yu,
  Joseph~E. Gonzalez, Hao Zhang, and Ion Stoica. 2023.
\newblock Efficient memory management for large language model serving with
  pagedattention.
\newblock In \emph{Proceedings of the ACM SIGOPS 29th Symposium on Operating
  Systems Principles}.

\bibitem[{Li et~al.(2024)Li, Sun, Yuan, Fan, hai zhao, and
  Liu}]{li2024generative}
Junlong Li, Shichao Sun, Weizhe Yuan, Run-Ze Fan, hai zhao, and Pengfei Liu.
  2024.
\newblock \href {https://openreview.net/forum?id=gtkFw6sZGS} {Generative judge
  for evaluating alignment}.
\newblock In \emph{The Twelfth International Conference on Learning
  Representations}.

\bibitem[{Ling et~al.(2017)Ling, Yogatama, Dyer, and
  Blunsom}]{ling-etal-2017-program}
Wang Ling, Dani Yogatama, Chris Dyer, and Phil Blunsom. 2017.
\newblock \href {https://doi.org/10.18653/v1/P17-1015} {Program induction by
  rationale generation: Learning to solve and explain algebraic word problems}.
\newblock In \emph{Proceedings of the 55th Annual Meeting of the Association
  for Computational Linguistics (Volume 1: Long Papers)}, pages 158--167,
  Vancouver, Canada. Association for Computational Linguistics.

\bibitem[{Liu et~al.(2024)Liu, Moosavi, and
  Lin}]{liu2024llmsnarcissisticevaluatorsego}
Yiqi Liu, Nafise~Sadat Moosavi, and Chenghua Lin. 2024.
\newblock \href {http://arxiv.org/abs/2311.09766} {Llms as narcissistic
  evaluators: When ego inflates evaluation scores}.

\bibitem[{Liusie et~al.(2024)Liusie, Manakul, and Gales}]{liusie-etal-2024-llm}
Adian Liusie, Potsawee Manakul, and Mark Gales. 2024.
\newblock \href {https://aclanthology.org/2024.eacl-long.8} {{LLM} comparative
  assessment: Zero-shot {NLG} evaluation through pairwise comparisons using
  large language models}.
\newblock In \emph{Proceedings of the 18th Conference of the European Chapter
  of the Association for Computational Linguistics (Volume 1: Long Papers)},
  pages 139--151, St. Julian{'}s, Malta. Association for Computational
  Linguistics.

\bibitem[{Oh et~al.(2024)Oh, Kim, Cha, and Oh}]{oh-etal-2024-generative}
Juhyun Oh, Eunsu Kim, Inha Cha, and Alice Oh. 2024.
\newblock \href {https://aclanthology.org/2024.eacl-srw.19} {The generative
  {AI} paradox in evaluation: {``}what it can solve, it may not evaluate{''}}.
\newblock In \emph{Proceedings of the 18th Conference of the European Chapter
  of the Association for Computational Linguistics: Student Research Workshop},
  pages 248--257, St. Julian{'}s, Malta. Association for Computational
  Linguistics.

\bibitem[{Panickssery et~al.(2024)Panickssery, Bowman, and
  Feng}]{panickssery2024llmevaluatorsrecognizefavor}
Arjun Panickssery, Samuel~R. Bowman, and Shi Feng. 2024.
\newblock \href {http://arxiv.org/abs/2404.13076} {Llm evaluators recognize and
  favor their own generations}.

\bibitem[{Pedregosa et~al.(2011)Pedregosa, Varoquaux, Gramfort, Michel,
  Thirion, Grisel, Blondel, Prettenhofer, Weiss, Dubourg, Vanderplas, Passos,
  Cournapeau, Brucher, Perrot, and Duchesnay}]{scikit-learn}
F.~Pedregosa, G.~Varoquaux, A.~Gramfort, V.~Michel, B.~Thirion, O.~Grisel,
  M.~Blondel, P.~Prettenhofer, R.~Weiss, V.~Dubourg, J.~Vanderplas, A.~Passos,
  D.~Cournapeau, M.~Brucher, M.~Perrot, and E.~Duchesnay. 2011.
\newblock Scikit-learn: Machine learning in {P}ython.
\newblock \emph{Journal of Machine Learning Research}, 12:2825--2830.

\bibitem[{Plank(2022)}]{plank-2022-problem}
Barbara Plank. 2022.
\newblock \href {https://doi.org/10.18653/v1/2022.emnlp-main.731} {The
  {``}problem{''} of human label variation: On ground truth in data, modeling
  and evaluation}.
\newblock In \emph{Proceedings of the 2022 Conference on Empirical Methods in
  Natural Language Processing}, pages 10671--10682, Abu Dhabi, United Arab
  Emirates. Association for Computational Linguistics.

\bibitem[{Raina et~al.(2024)Raina, Liusie, and Gales}]{raina2024llmasajudge}
Vyas Raina, Adian Liusie, and Mark Gales. 2024.
\newblock \href {http://arxiv.org/abs/2402.14016} {Is llm-as-a-judge robust?
  investigating universal adversarial attacks on zero-shot llm assessment}.

\bibitem[{Seabold and Perktold(2010)}]{seabold2010statsmodels}
Skipper Seabold and Josef Perktold. 2010.
\newblock statsmodels: Econometric and statistical modeling with python.
\newblock In \emph{9th Python in Science Conference}.

\bibitem[{Shaib et~al.(2024)Shaib, Elazar, Li, and
  Wallace}]{shaib-etal-2024-detection}
Chantal Shaib, Yanai Elazar, Junyi~Jessy Li, and Byron~C Wallace. 2024.
\newblock \href {https://doi.org/10.18653/v1/2024.emnlp-main.368} {Detection
  and measurement of syntactic templates in generated text}.
\newblock In \emph{Proceedings of the 2024 Conference on Empirical Methods in
  Natural Language Processing}, pages 6416--6431, Miami, Florida, USA.
  Association for Computational Linguistics.

\bibitem[{Tan et~al.(2024)Tan, Li, Wang, Beigi, Jiang, Bhattacharjee, Karami,
  Li, Cheng, and Liu}]{tan2024largelanguagemodelsdata}
Zhen Tan, Dawei Li, Song Wang, Alimohammad Beigi, Bohan Jiang, Amrita
  Bhattacharjee, Mansooreh Karami, Jundong Li, Lu~Cheng, and Huan Liu. 2024.
\newblock \href {http://arxiv.org/abs/2402.13446} {Large language models for
  data annotation: A survey}.

\bibitem[{Wang et~al.(2024{\natexlab{a}})Wang, Li, Chen, Cai, Zhu, Lin, Cao,
  Kong, Liu, Liu, and Sui}]{wang-etal-2024-large-language-models-fair}
Peiyi Wang, Lei Li, Liang Chen, Zefan Cai, Dawei Zhu, Binghuai Lin, Yunbo Cao,
  Lingpeng Kong, Qi~Liu, Tianyu Liu, and Zhifang Sui. 2024{\natexlab{a}}.
\newblock \href {https://doi.org/10.18653/v1/2024.acl-long.511} {Large language
  models are not fair evaluators}.
\newblock In \emph{Proceedings of the 62nd Annual Meeting of the Association
  for Computational Linguistics (Volume 1: Long Papers)}, pages 9440--9450,
  Bangkok, Thailand. Association for Computational Linguistics.

\bibitem[{Wang et~al.(2024{\natexlab{b}})Wang, Yu, Yao, Zeng, Yang, Wang, Chen,
  Jiang, Xie, Wang, Xie, Ye, Zhang, and Zhang}]{wang2024pandalm}
Yidong Wang, Zhuohao Yu, Wenjin Yao, Zhengran Zeng, Linyi Yang, Cunxiang Wang,
  Hao Chen, Chaoya Jiang, Rui Xie, Jindong Wang, Xing Xie, Wei Ye, Shikun
  Zhang, and Yue Zhang. 2024{\natexlab{b}}.
\newblock \href {https://openreview.net/forum?id=5Nn2BLV7SB} {Panda{LM}: An
  automatic evaluation benchmark for {LLM} instruction tuning optimization}.
\newblock In \emph{The Twelfth International Conference on Learning
  Representations}.

\bibitem[{Wang et~al.(2024{\natexlab{c}})Wang, Dong, Delalleau, Zeng, Shen,
  Egert, Zhang, Sreedhar, and Kuchaiev}]{wang2024helpsteer2}
Zhilin Wang, Yi~Dong, Olivier Delalleau, Jiaqi Zeng, Gerald Shen, Daniel Egert,
  Jimmy~J. Zhang, Makesh~Narsimhan Sreedhar, and Oleksii Kuchaiev.
  2024{\natexlab{c}}.
\newblock \href {http://arxiv.org/abs/2406.08673} {Helpsteer2: Open-source
  dataset for training top-performing reward models}.

\bibitem[{Wei et~al.(2022)Wei, Wang, Schuurmans, Bosma, brian ichter, Xia, Chi,
  Le, and Zhou}]{wei2022chain}
Jason Wei, Xuezhi Wang, Dale Schuurmans, Maarten Bosma, brian ichter, Fei Xia,
  Ed~H. Chi, Quoc~V Le, and Denny Zhou. 2022.
\newblock \href {https://openreview.net/forum?id=_VjQlMeSB_J} {Chain of thought
  prompting elicits reasoning in large language models}.
\newblock In \emph{Advances in Neural Information Processing Systems}.

\bibitem[{West et~al.(2024)West, Lu, Dziri, Brahman, Li, Hwang, Jiang, Fisher,
  Ravichander, Chandu, Newman, Koh, Ettinger, and Choi}]{west2024the}
Peter West, Ximing Lu, Nouha Dziri, Faeze Brahman, Linjie Li, Jena~D. Hwang,
  Liwei Jiang, Jillian Fisher, Abhilasha Ravichander, Khyathi Chandu, Benjamin
  Newman, Pang~Wei Koh, Allyson Ettinger, and Yejin Choi. 2024.
\newblock \href {https://openreview.net/forum?id=CF8H8MS5P8} {The generative
  {AI} paradox: {\textquotedblleft}what it can create, it may not
  understand{\textquotedblright}}.
\newblock In \emph{The Twelfth International Conference on Learning
  Representations}.

\bibitem[{Wolf et~al.(2020)Wolf, Debut, Sanh, Chaumond, Delangue, Moi, Cistac,
  Rault, Louf, Funtowicz, Davison, Shleifer, von Platen, Ma, Jernite, Plu, Xu,
  Le~Scao, Gugger, Drame, Lhoest, and Rush}]{wolf-etal-2020-transformers}
Thomas Wolf, Lysandre Debut, Victor Sanh, Julien Chaumond, Clement Delangue,
  Anthony Moi, Pierric Cistac, Tim Rault, Remi Louf, Morgan Funtowicz, Joe
  Davison, Sam Shleifer, Patrick von Platen, Clara Ma, Yacine Jernite, Julien
  Plu, Canwen Xu, Teven Le~Scao, Sylvain Gugger, Mariama Drame, Quentin Lhoest,
  and Alexander Rush. 2020.
\newblock \href {https://doi.org/10.18653/v1/2020.emnlp-demos.6} {Transformers:
  State-of-the-art natural language processing}.
\newblock In \emph{Proceedings of the 2020 Conference on Empirical Methods in
  Natural Language Processing: System Demonstrations}, pages 38--45, Online.
  Association for Computational Linguistics.

\bibitem[{Wu and Aji(2025)}]{wu-aji-2025-style}
Minghao Wu and Alham~Fikri Aji. 2025.
\newblock \href {https://aclanthology.org/2025.coling-main.21/} {Style over
  substance: Evaluation biases for large language models}.
\newblock In \emph{Proceedings of the 31st International Conference on
  Computational Linguistics}, pages 297--312, Abu Dhabi, UAE. Association for
  Computational Linguistics.

\bibitem[{Wu et~al.(2024)Wu, Yuan, Golovneva, Xu, Tian, Jiao, Weston, and
  Sukhbaatar}]{wu2024metarewardinglanguagemodelsselfimproving}
Tianhao Wu, Weizhe Yuan, Olga Golovneva, Jing Xu, Yuandong Tian, Jiantao Jiao,
  Jason Weston, and Sainbayar Sukhbaatar. 2024.
\newblock \href {http://arxiv.org/abs/2407.19594} {Meta-rewarding language
  models: Self-improving alignment with llm-as-a-meta-judge}.

\bibitem[{Xu et~al.(2024)Xu, Zhu, Zhao, Pan, Li, and
  Wang}]{xu2024prideprejudicellmamplifies}
Wenda Xu, Guanglei Zhu, Xuandong Zhao, Liangming Pan, Lei Li, and William~Yang
  Wang. 2024.
\newblock \href {http://arxiv.org/abs/2402.11436} {Pride and prejudice: Llm
  amplifies self-bias in self-refinement}.

\bibitem[{Yang et~al.(2024)Yang, Yang, Hui, Zheng, Yu, Zhou, Li, Li, Liu,
  Huang, Dong, Wei, Lin, Tang, Wang, Yang, Tu, Zhang, Ma, Xu, Zhou, Bai, He,
  Lin, Dang, Lu, Chen, Yang, Li, Xue, Ni, Zhang, Wang, Peng, Men, Gao, Lin,
  Wang, Bai, Tan, Zhu, Li, Liu, Ge, Deng, Zhou, Ren, Zhang, Wei, Ren, Fan, Yao,
  Zhang, Wan, Chu, Liu, Cui, Zhang, and Fan}]{qwen2}
An~Yang, Baosong Yang, Binyuan Hui, Bo~Zheng, Bowen Yu, Chang Zhou, Chengpeng
  Li, Chengyuan Li, Dayiheng Liu, Fei Huang, Guanting Dong, Haoran Wei, Huan
  Lin, Jialong Tang, Jialin Wang, Jian Yang, Jianhong Tu, Jianwei Zhang,
  Jianxin Ma, Jin Xu, Jingren Zhou, Jinze Bai, Jinzheng He, Junyang Lin, Kai
  Dang, Keming Lu, Keqin Chen, Kexin Yang, Mei Li, Mingfeng Xue, Na~Ni, Pei
  Zhang, Peng Wang, Ru~Peng, Rui Men, Ruize Gao, Runji Lin, Shijie Wang, Shuai
  Bai, Sinan Tan, Tianhang Zhu, Tianhao Li, Tianyu Liu, Wenbin Ge, Xiaodong
  Deng, Xiaohuan Zhou, Xingzhang Ren, Xinyu Zhang, Xipin Wei, Xuancheng Ren,
  Yang Fan, Yang Yao, Yichang Zhang, Yu~Wan, Yunfei Chu, Yuqiong Liu, Zeyu Cui,
  Zhenru Zhang, and Zhihao Fan. 2024.
\newblock Qwen2 technical report.
\newblock \emph{arXiv preprint arXiv:2407.10671}.

\bibitem[{Young et~al.(2024)Young, Chen, Li, Huang, Zhang, Zhang, Li, Zhu,
  Chen, Chang, Yu, Liu, Liu, Yue, Yang, Yang, Yu, Xie, Huang, Hu, Ren, Niu,
  Nie, Xu, Liu, Wang, Cai, Gu, Liu, and Dai}]{ai2024yi}
Alex Young, Bei Chen, Chao Li, Chengen Huang, Ge~Zhang, Guanwei Zhang, Heng Li,
  Jiangcheng Zhu, Jianqun Chen, Jing Chang, Kaidong Yu, Peng Liu, Qiang Liu,
  Shawn Yue, Senbin Yang, Shiming Yang, Tao Yu, Wen Xie, Wenhao Huang, Xiaohui
  Hu, Xiaoyi Ren, Xinyao Niu, Pengcheng Nie, Yuchi Xu, Yudong Liu, Yue Wang,
  Yuxuan Cai, Zhenyu Gu, Zhiyuan Liu, and Zonghong Dai. 2024.
\newblock \href {http://arxiv.org/abs/2403.04652} {Yi: Open foundation models
  by 01.ai}.

\bibitem[{Yuan et~al.(2024)Yuan, Pang, Cho, Li, Sukhbaatar, Xu, and
  Weston}]{yuan2024selfrewardinglanguagemodels}
Weizhe Yuan, Richard~Yuanzhe Pang, Kyunghyun Cho, Xian Li, Sainbayar
  Sukhbaatar, Jing Xu, and Jason Weston. 2024.
\newblock \href {http://arxiv.org/abs/2401.10020} {Self-rewarding language
  models}.

\bibitem[{Zheng et~al.(2023)Zheng, Chiang, Sheng, Zhuang, Wu, Zhuang, Lin, Li,
  Li, Xing, Zhang, Gonzalez, and Stoica}]{zheng2023judging}
Lianmin Zheng, Wei-Lin Chiang, Ying Sheng, Siyuan Zhuang, Zhanghao Wu, Yonghao
  Zhuang, Zi~Lin, Zhuohan Li, Dacheng Li, Eric Xing, Hao Zhang, Joseph~E.
  Gonzalez, and Ion Stoica. 2023.
\newblock \href {https://openreview.net/forum?id=uccHPGDlao} {Judging
  {LLM}-as-a-judge with {MT}-bench and chatbot arena}.
\newblock In \emph{Thirty-seventh Conference on Neural Information Processing
  Systems Datasets and Benchmarks Track}.

\end{thebibliography}
\bibliographystyle{acl_natbib}

\appendix

\section{Experimental Setup}
\label{app_sec:experimental_setup}

We provide further details on the general setup described in Section \ref{sec:setup}. Specifically, we include statistics and examples of the datasets, additional information on the models used, and the exact prompts employed in this study.

\subsection{Datasets}
\label{app_sec:datasets}

Additional information about the datasets is given in Table \ref{tab:dataset_statistics}, which presents an overview of the dataset statistics. Note that for the MATH dataset, we only include the most challenging questions, called levels 4 and 5, in the dataset.
Notably, it has ground truth answer sequences that are, on average, almost three times longer than those in other datasets.
\\
\\
In Table \ref{app_tab:example_few_shots}, we provide examples of questions and their corresponding answers from the ground truth. Note that these examples were used for few-shot prompting.

\begin{table}
\centering
\resizebox{.47\textwidth}{!}{
\begin{tabular}{lccc}
\toprule
 & \multirow{2}{*}{\# questions} & \multicolumn{1}{c}{Avg.} & \multicolumn{1}{c}{Avg.} \\
 &  & \# question characters & \# answer characters \\
\midrule
AQUA-RAT & 254 & 239.1 & 203.1 \\
MATH & 1516 & 216.5 & 643.9 \\
GSM8K & 1319 & 239.9 & 292.9 \\
\bottomrule
\end{tabular}
}
\caption{An overview of dataset size and text length.}
\label{tab:dataset_statistics}
\end{table}

\begin{table*}
\centering
\resizebox{.95\textwidth}{!}{
\begin{tabular}{l|p{6cm}|p{6cm}}
\toprule
 & Question & Answer \\
\midrule
AQUA-RAT & Two friends plan to walk along a 43-km trail, starting at opposite ends of the trail at the same time. If Friend P's rate is 15\% faster than Friend Q's, how many kilometers will Friend P have walked when they pass each other?
Options:
A)21
B)21.5
C)22
D)22.5
E)23 & If Q complete x kilometers, then P completes 1.15x kilometers.
x + 1.15x = 43
2.15x=43
x = 43/2.15 = 20
Then P will have have walked 1.15*20=23 km.
The answer is E. \#\#\#\# E \\
\hline
GSM8K & Natalia sold clips to 48 of her friends in April, and then she sold half as many clips in May. How many clips did Natalia sell altogether in April and May? & Natalia sold 48/2 = <<48/2=24>>24 clips in May.
Natalia sold 48+24 = <<48+24=72>>72 clips altogether in April and May.
\#\#\#\# 72 \\
\hline
MATH & Mr. Madoff invests 1000 dollars in a fund that compounds annually at a constant interest rate.  After three years, his investment has grown to 1225 dollars.  What is the annual interest rate, as a percentage?  (Round your answer to the nearest integer.) & Let $r$ be the annual interest rate.  Then after three years, Mr. Madoff's investment is $1000 \cdot \left( 1 + \frac{r}{100} \right)^3 $, so $ 1000 \cdot \left( 1 + \frac{r}{100} \right)^3 = 1225.$ Then $\left( 1 + \frac{r}{100} \right)^3 = 1.225,$so $[1 + \frac{r}{100} = \sqrt[3]{1.225} = 1.069987 \dots,$ which means $r = \boxed{7}$, to the nearest integer. \#\#\#\#  7.0 \\
\bottomrule
\end{tabular}
}
\caption{Example of ground truth answers used for few-shot prompting.}
\label{app_tab:example_few_shots}
\end{table*}

\subsection{Models}
\label{app_subsec:models}

We execute all models using the VLLM software for LLM serving \cite{kwon2023efficient}. The weights for all models are accessible through Huggingface Transformers \cite{wolf-etal-2020-transformers}. Table \ref{tab:model_links} includes hyperlinks to each model for easy reference.

\begin{table}
    \centering
\resizebox{.47\textwidth}{!}{
\begin{tabular}{l|l}
\toprule
Model & URL \\
\midrule
Llama 3.1 70B & \url{https://huggingface.co/meta-llama/Llama-3.1-70B-Instruct} \\
Qwen 2.5 72B & \url{https://huggingface.co/Qwen/Qwen2.5-72B-Instruct} \\
Qwen 2.5 14B & \url{https://huggingface.co/Qwen/Qwen2.5-14B-Instruct} \\
Gemma 2 27B & \url{https://huggingface.co/google/gemma-2-27b-it} \\
Qwen 2.5 7B & \url{https://huggingface.co/Qwen/Qwen2.5-7B-Instruct} \\
Gemma 2 9B & \url{https://huggingface.co/google/gemma-1.1-9b-it} \\
Llama 3.1 8B & \url{https://huggingface.co/meta-llama/Llama-3.1-8B-Instruct} \\
Gemma 2 2B & \url{https://huggingface.co/google/gemma-1.1-2b-it} \\
\bottomrule
\end{tabular}
}
\caption{Used models and corresponding hyperlinks.}
\label{tab:model_links}
\end{table}

\subsection{Prompts}
\label{app_sec:prompts}

\begin{figure}
\centering
\begin{tcolorbox}[ title=User, colframe=black!10, coltitle=black, fonttitle=\bfseries, boxrule=0.5mm, width=\columnwidth, fontupper=\small]
You are a reasoning assistant. Always answer exactly in the same format. Use '\#\#\#\#' to separate the final answer (without additional comments) from the reasoning. 
\\
\\
<< Few-Shot Question 1 >>
\end{tcolorbox}
\begin{tcolorbox}[ title=Assistant, colframe=black!10, coltitle=black, fonttitle=\bfseries, boxrule=0.5mm, width=\columnwidth, fontupper=\small]
<< Few-Shot Answer 1 >>
\end{tcolorbox}

\begin{tcolorbox}[ title=..., colframe=black!10, coltitle=black, fonttitle=\bfseries, boxrule=0.5mm, width=\columnwidth, fontupper=\small]
...
\end{tcolorbox}
\begin{tcolorbox}[ title=User, colframe=black!10, coltitle=black, fonttitle=\bfseries, boxrule=0.5mm, width=\columnwidth, fontupper=\small]
<< Few-Shot Question 4 >>
\end{tcolorbox}
\begin{tcolorbox}[ title=Assistant, colframe=black!10, coltitle=black, fonttitle=\bfseries, boxrule=0.5mm, width=\columnwidth, fontupper=\small]
<< Few-Shot Answer 4 >>
\end{tcolorbox}
\begin{tcolorbox}[ title=User, colframe=black!10, coltitle=black, fonttitle=\bfseries, boxrule=0.5mm, width=\columnwidth, fontupper=\small]
<< Sample Question >>
\end{tcolorbox}
\caption{The prompt to solve tasks. Few-shots and actual questions are filled in within ``<<'' and ``>>'' symbols. }
\label{app_fig:initial_prompt}
\end{figure} 
\begin{figure}
\centering
\begin{tcolorbox}[title=User, colframe=black!10, coltitle=black, fonttitle=\bfseries, boxrule=0.5mm, width=\columnwidth, fontupper=\small]
Question: \\
<< question >>  \\
\\
Answer A: \\
<< answer A >>  \\
-------------- \\
Answer B: \\
<< answer B >>  \\
-------------- \\
 \\
Compare both answers in detail and decide whether both answers are correct, both answers are incorrect or whether answer 1 or answer 2 is correct.
\\
\\
Conclude with a JSON in Markdown format indicating your choice between "answer\_1", "answer\_2", "both\_correct" or "both\_incorrect":
{```}json  \\
\{ \\
$ $ $ $ "answer":  "..." \\
\} \\
{```} \\
\end{tcolorbox}

\caption{Judge Prompt. Candidate answers are filled in within ``<<'' and ``>>'' symbols. }
\label{app_fig:judge_prompt}
\end{figure}

We used two different prompts within this project. In general, we designed the prompts to be minimial, by assigning a minimal personality, a quick task description, and description of the output format.
The prompt shown in Figure \ref{app_fig:initial_prompt} is used for the candidate solution generation for all datasets. Examples of the few-shots are in Table \ref{app_tab:example_few_shots}.
The prompt for the judges is given in Figure \ref{app_fig:judge_prompt}. Note that we run experiments for both orders of the answers of the models $A$ and $B$.

\subsection{Infrastructure}

The experiments were run on NVIDIA A100 and NVIDIA H100. The judgments used in Section \ref{sec:general_performance} took around 3 day equivalents on 4 A100 40GB. Using 2 H100 90GB and 4 A100 40 GB it took less than 2 days.

\section{General Performance}
\label{app_sec:general_performance}

This section provides additional information related to Section \ref{sec:general_performance}. Specifically, we present the task performance of all models across all datasets, as well as the judging performance of all models when used as judges.

\subsection{Task Performance}
\label{app_subsec:task_performance}

\begin{table}
\centering
\resizebox{.47\textwidth}{!}{
\begin{tabular}{llll}
\toprule
 & GSM8K & AQUA-RAT & MATH \\
\midrule
Llama 3.1 70B & 93.25 & 78.57 & 47.11 \\
Qwen 2.5 72B & \textbf{95.07} & \textbf{83.73} & \textbf{73.86} \\
Qwen 2.5 14B & \underline{93.48} & \underline{82.54} & \underline{64.47} \\
Gemma 2 27B & 85.97 & 67.46 & 38.80 \\
Qwen 2.5 7B & 88.10 & 75.40 & 60.31 \\
Gemma 2 9B & 80.52 & 61.51 & 31.31 \\
Llama 3.1 8B & 72.40 & 61.51 & 20.69 \\
Gemma 2 2B & 37.53 & 26.98 & 7.15 \\
\bottomrule
\end{tabular}
}
\caption{Task performance of the investigated models.}
\label{tab:initial_model_performance_per_dataset}
\end{table}

In various contexts in this work, the task performance of the individual models is essential. Therefore, we provide the accuracy of all models and all datasets in Table \ref{tab:initial_model_performance_per_dataset}.

\subsection{Judging performance per model pair}

\begin{figure}
\centering
\begin{subfigure}{0.228\textwidth}
    \centering
    \includegraphics[width=\textwidth]{figures/performance_per_pair/p_j_given_a_b/meta-llama_Llama-3.1-70B-Instruct.png}
    \caption{LLama 3.1 70B}
\end{subfigure}
\begin{subfigure}{0.235\textwidth}
    \centering
    \includegraphics[width=\textwidth]{figures/performance_per_pair/p_j_given_a_b/Qwen_Qwen2.5-72B-Instruct.png}
    \caption{Qwen 2.5 72B}
\end{subfigure}
\begin{subfigure}{0.228\textwidth}
    \centering
    \includegraphics[width=\textwidth]{figures/performance_per_pair/p_j_given_a_b/Qwen_Qwen2.5-14B-Instruct.png}
    \caption{Qwen 2.5 14B}
\end{subfigure}
\begin{subfigure}{0.235\textwidth}
    \centering
    \includegraphics[width=\textwidth]{figures/performance_per_pair/p_j_given_a_b/google_gemma-2-27b-it.png}
    \caption{Gemma 2 27B}
\end{subfigure}


\begin{subfigure}{0.228\textwidth}
    \centering
    \includegraphics[width=\textwidth]{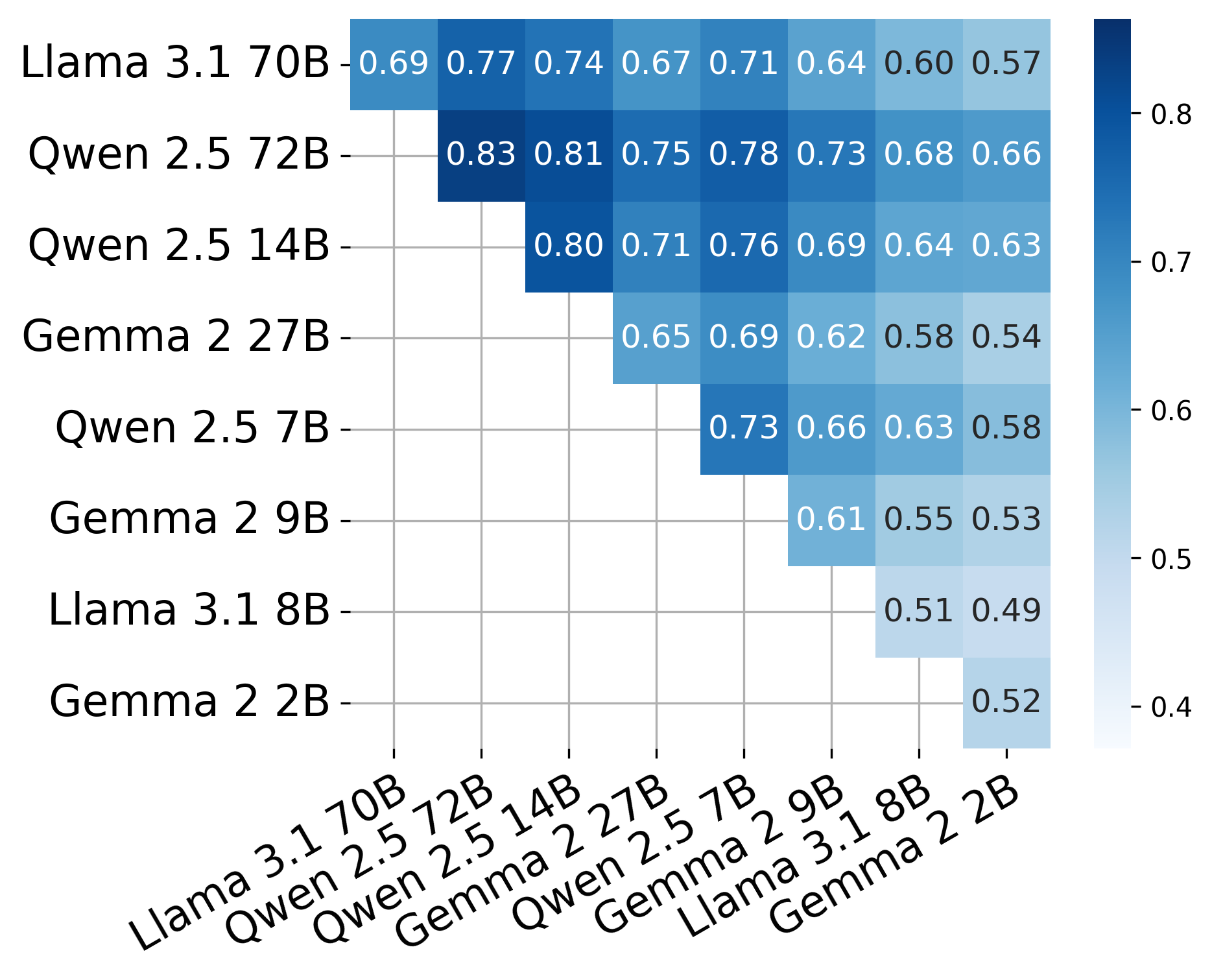}
    \caption{Qwen 2.5 7B}
\end{subfigure}
\begin{subfigure}{0.235\textwidth}
    \centering
    \includegraphics[width=\textwidth]{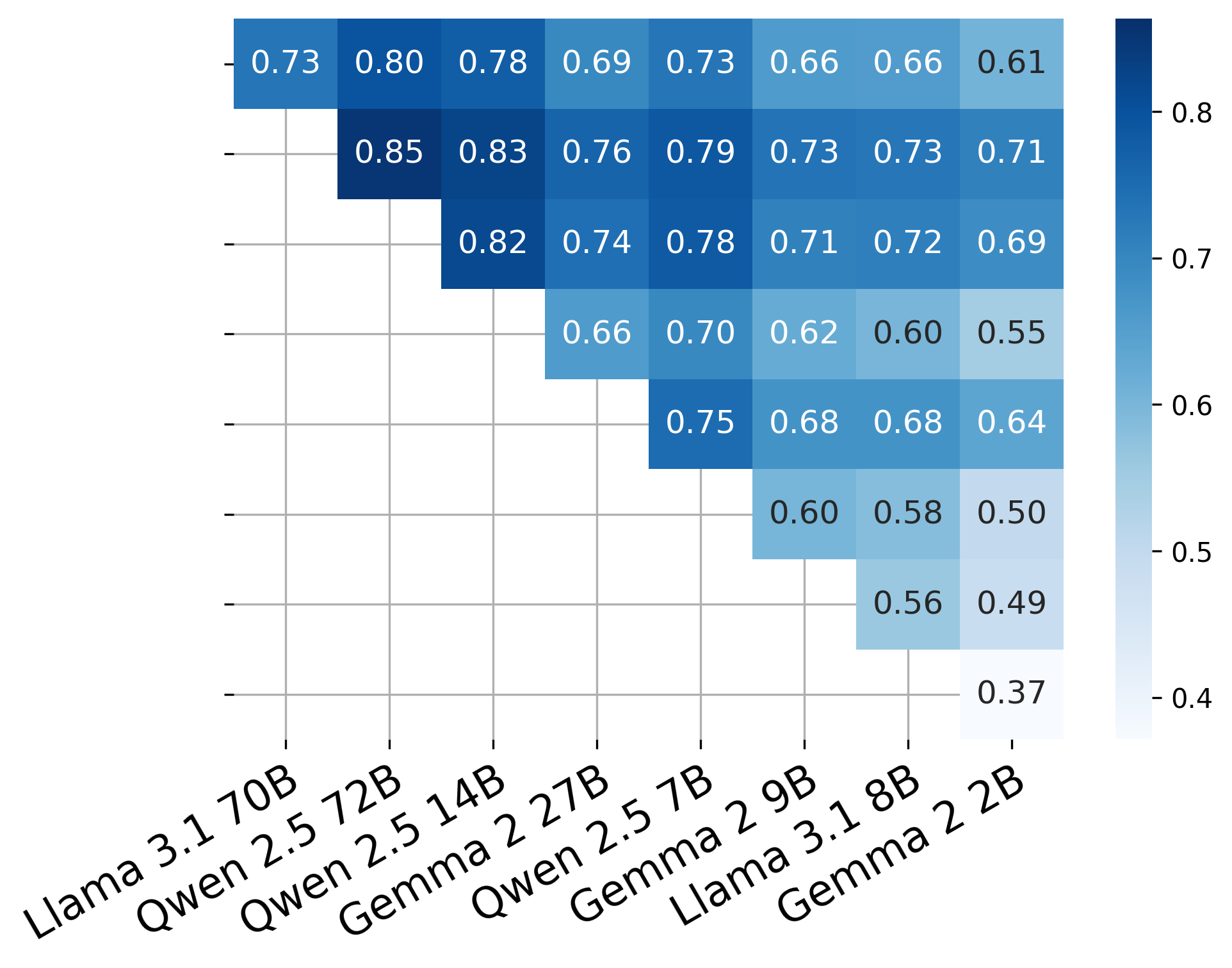}
    \caption{Gemma 2 9B}
\end{subfigure}
\begin{subfigure}{0.228\textwidth}
    \centering
    \includegraphics[width=\textwidth]{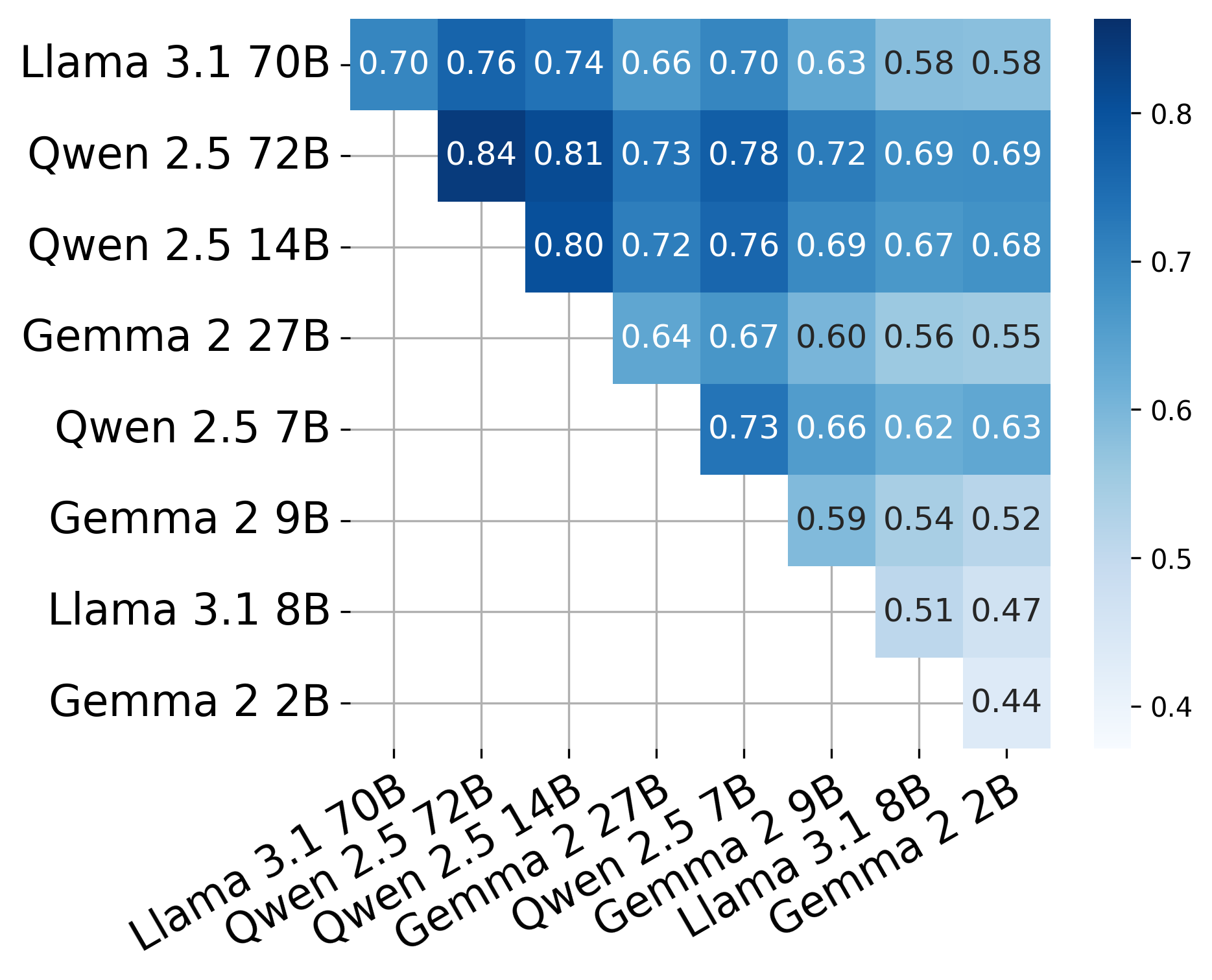}
    \caption{LLama 3.1 8B}
\end{subfigure}
\begin{subfigure}{0.235\textwidth}
    \centering
    \includegraphics[width=\textwidth]{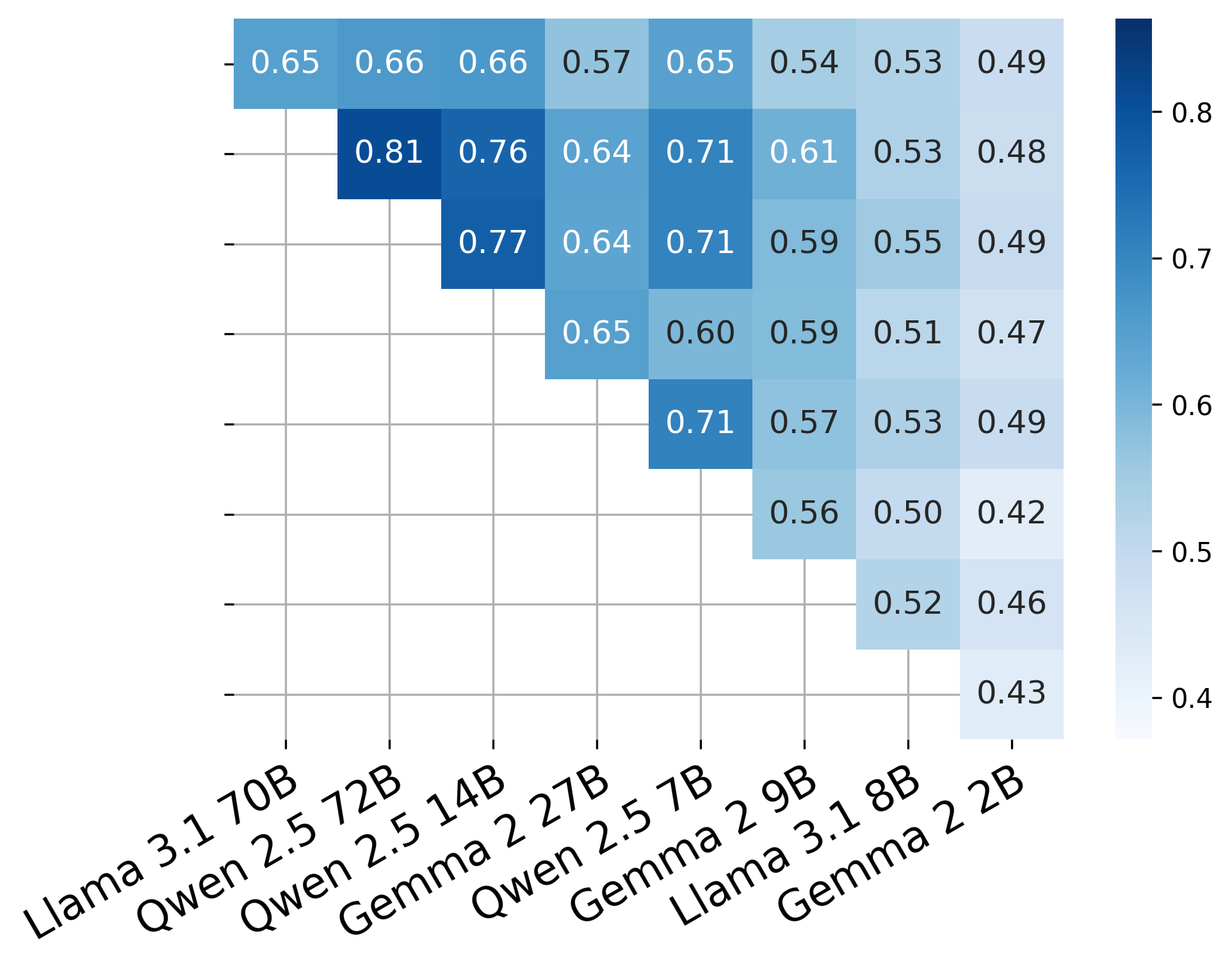}
    \caption{Gemma 2 2B}
\end{subfigure}

\caption{Performance $S^J_{A,B}$ of LLM judges on model pairs, averaged across datasets.}
\label{tab:performance_per_pair_full}
\end{figure}

We conduct experiments with all eight models serving as judges. We present the performance metrics of all judges across all model pairs in Figure \ref{tab:performance_per_pair_full}.

\section{Examples}

\subsection{Example Subset Performance}
\label{app_sec:subset}

\begin{table*}
\centering
\resizebox{.96\textwidth}{!}{
\begin{tabular}{llllrrrr}
\toprule
Judge & model A & model B & dataset & $S_A$ & $S_B$ & $S^{J}_{A,B}$ & No. Samples \\
\midrule
Llama 3.1 70B & Qwen 2.5 14B & Gemma 2 2B & MATH & 64.50 & 7.10 & 94.60 & 1655 \\
Llama 3.1 70B & Qwen 2.5 72B & Gemma 2 2B & AQUA-RAT & 83.70 & 27.00 & 94.50 & 309 \\
Llama 3.1 70B & Qwen 2.5 7B & Gemma 2 2B & MATH & 60.30 & 7.10 & 94.00 & 1520 \\
Llama 3.1 70B & Qwen 2.5 72B & Qwen 2.5 7B & AQUA-RAT & 83.70 & 75.40 & 62.50 & 64 \\
Llama 3.1 70B & Qwen 2.5 7B & Qwen 2.5 14B & AQUA-RAT & 75.40 & 82.50 & 42.30 & 26 \\
Llama 3.1 70B & Qwen 2.5 72B & Qwen 2.5 72B & MATH & 73.90 & 73.90 & 49.50 & 206 \\
Llama 3.1 70B & Gemma 2 27B & Qwen 2.5 14B & AQUA-RAT & 67.50 & 82.50 & 15.00 & 20 \\
Llama 3.1 70B & Gemma 2 2B & Qwen 2.5 14B & GSM8K & 37.50 & 93.50 & 15.00 & 20 \\
Llama 3.1 70B & Gemma 2 2B & Llama 3.1 70B & MATH & 7.10 & 47.10 & 14.50 & 62 \\
\bottomrule
\end{tabular}
}
\caption{Examples of judgement performances on subsets where model $A$ is correct and model $B$ is incorrect.}
\label{tab:subset_examples}
\end{table*}

To better understand the correlation observed in Figure \ref{fig:performance_diff_vs_accuracy}, we provide examples of these subsets, which can be seen in Table \ref{tab:subset_examples}. 
These examples include the following details: the judge, the compared models, the dataset, the performance of the correct model on the dataset (denoted by $S_A$), the performance of the incorrect model on the dataset $S_B$, the judgment performance on the subset (denoted by $S^{J}_{A,B}$), and the size of the subset. We provide the three subsets with the highest performance, the three subsets with the lowest performance, and three random subsets where Llama 3.1 70B is the judge.

\section{Statistical Methodology}
\label{a:stats}

We describe the statistical background for the tests applied in Section \ref{sec:prediction}. All predictions and statistical tests in Section \ref{sec:prediction} were performed using the statsmodels library \cite{seabold2010statsmodels}.

\subsection{Coefficient of Determination}

The coefficient of determination, $R^2$, for evaluation of linear regression models \citep{fahrmeir2013regression} is defined as follows:

$$R^2 = \frac{\sum_{i=1}^{n} (\hat y_i - \bar y)^2}{\sum_{i=1}^{n} (y_i - \bar y)^2}$$

$R^2$ measures the share of the variance in $Y$ explained by its covariation with the features $\mathbf{X}$ included in the model by dividing the variation of the \textit{predicted} values $\hat y_i$ by the variation of the true target values $y_i$. If the features $\mathbf{X}$ have high explanatory power for $Y$, the $\hat y_i$ will be close to the $y_i$ and $R^2$ will be close to 1, while in the extreme case of no correlation between $\mathbf{X}$ and $Y$ the arithmetic mean is the best estimate (i.e., $\hat y_i = \bar y\; \forall\; i = 1, \dots, n$) resulting in $R^2 = 0$.

\subsection{Overall-F-Test}

The Overall-F-Test is built upon $R^2$ and tests whether the overall model is of any significant value for explaining the variation of the target variable. The F-distributed test statistic is calculated as

$$\frac{R^2}{1 - R^2} \cdot \frac{n - p - 1}{p},$$

where $R^2$ is the coefficient of determination, $n$ is the number of observations, and $p$ is the number of covariates included in the model (i.e., the number of estimated coefficients excluding the intercept). The hypotheses that can be tested this way are

$$H_0: \beta_1 = \beta_2 = \dots = \beta_p = 0$$ vs. $$\qquad H_1: \beta_j \neq 0\; \mbox{for at least one}\; j \in \{1, \dots, p\}.$$

So from a rejection of $H_0$, it can be concluded that at least one of the included features exhibits explanatory power for the variation of the target variable.

\subsection{Multiple Testing}

Since we conduct multiple statistical tests within the scope of one research project, it is important to consider multiple testing as a potential problem resulting in false positive findings. The p-values from our tests, however, also satisfy a significance level resulting from a Bonferroni Correction of the typical significance level of 5\%. 

\section{Sample-level Analysis}
\label{app_sec:sample_level}

We utilize Scikit-learn \cite{scikit-learn} library to train and evaluate the Logistic Regression and Random Forest Model. We use the standard settings for the Logistic Regression model. We use the Random Forests model with 1500 estimators, and standard settings apart from that.
\\
\\
During preprocessing, we use simple word splittling by spaces. We employ the english stop word removal integreated into Scikit-learn. We set the maximum number of features to 5,000, for the N-Gram of part-of-speech tags, we set the N-gram range from 5-grams up to 13-grams, following settings of \citet{shaib-etal-2024-detection}. For training, we use the Scikit-learn \cite{scikit-learn} library. The running time was negligible.

In Table \ref{tab:prediction_stats} 

\begin{table}
\centering
\resizebox{.49\textwidth}{!}{
\begin{tabular}{lcccc}
\toprule
Model & Both correct & A correct & B correct & Both incorrect \\
\midrule
Llama 3.1 70B & 51.8 & 18.1 & 21.7 & 8.4 \\
Qwen 2.5 72B & 54.9 & 19.6 & 19.8 & 5.6 \\
Qwen 2.5 14B & 50.2 & 20.0 & 23.0 & 6.9 \\
Gemma 2 27B & 52.6 & 15.9 & 15.4 & 16.0 \\
\bottomrule
\end{tabular}
}
\caption{Percentage of predictions individual models made.}
\label{tab:prediction_stats}
\end{table}

\end{document}